\documentclass{article}

     \PassOptionsToPackage{numbers, compress}{natbib}

\usepackage[preprint]{neurips_2022}

\usepackage[utf8]{inputenc} 
\usepackage[T1]{fontenc}    
\usepackage[hidelinks]{hyperref}       
\usepackage{url}            
\usepackage{booktabs}       
\usepackage{amsfonts}       
\usepackage{nicefrac}       
\usepackage{microtype}      
\usepackage{xcolor}         

\usepackage{times}
\usepackage{latexsym}

\usepackage[utf8]{inputenc} 
\usepackage[T1]{fontenc}    
\usepackage{url}            
\usepackage{booktabs}       
\usepackage{amsfonts}       
\usepackage{nicefrac}       
\usepackage{microtype}      
\usepackage{amsmath,amssymb}
\usepackage{epsfig,graphicx,subfigure,caption}
\usepackage{algpseudocode}
\usepackage{multirow,booktabs, hhline}
\usepackage[ruled,noend]{algorithm2e}
\usepackage{amsmath, bm}
\usepackage{titlesec}
\usepackage{wrapfig}
\usepackage{lipsum}
\usepackage{caption}
\usepackage{footnote}
\usepackage[bottom]{footmisc}
\usepackage[toc,page]{appendix}
\usepackage{todonotes}

\newcommand{\baby}{X$^2$-VLM\xspace}
\newcommand{\babyx}{X$^2$-VLM}
\newcommand{\babyB}{X$^2$-VLM$_\mathrm{base}$\xspace}
\newcommand{\babyL}{X$^2$-VLM$_\mathrm{large}$\xspace}

\title{X$^2$-VLM: All-In-One Pre-trained Model For Vision-Language Tasks}

\author{
Yan Zeng\thanks{Correspondence to: <zengyan.yanne@bytedance.com>.} \\
ByteDance Research \\
\And
Xinsong Zhang \\
ByteDance Research \\
\And
Hang Li \\
ByteDance Research \\
\AND 
Jiawei Wang \\
ByteDance Research \\
\And
Jipeng Zhang \\
HKUST \\
\And
Wangchunshu Zhou \\
ETH Zurich
}

\begin{document}
\maketitle

\begin{abstract}
Vision language pre-training aims to learn alignments between vision and language from a large amount of data. Most existing methods only learn image-text alignments. Some others utilize pre-trained object detectors to leverage vision language alignments at the object level. In this paper, we propose to learn multi-grained vision language alignments by a unified pre-training framework that learns multi-grained aligning and multi-grained localization simultaneously. Based on it, we present \baby, an all-in-one model with a flexible modular architecture, in which we further unify image-text pre-training and video-text pre-training in one model. \baby is able to learn unlimited visual concepts associated with diverse text descriptions. Experiment results show that \baby performs the best on base and large scale for both image-text and video-text tasks, making a good trade-off between performance and model scale. Moreover, we show that the modular design of \baby results in high transferability for it to be utilized in any language or domain. For example, by simply replacing the text encoder with XLM-R, \baby outperforms state-of-the-art multilingual multi-modal pre-trained models without any multilingual pre-training. The code and pre-trained models are available at \url{github.com/zengyan-97/X2-VLM}. 
\end{abstract}

\section{Introduction}
\label{sec:introduction}

\begin{figure*}[ht]
\begin{center}
\centerline{\includegraphics[width=0.9\columnwidth]{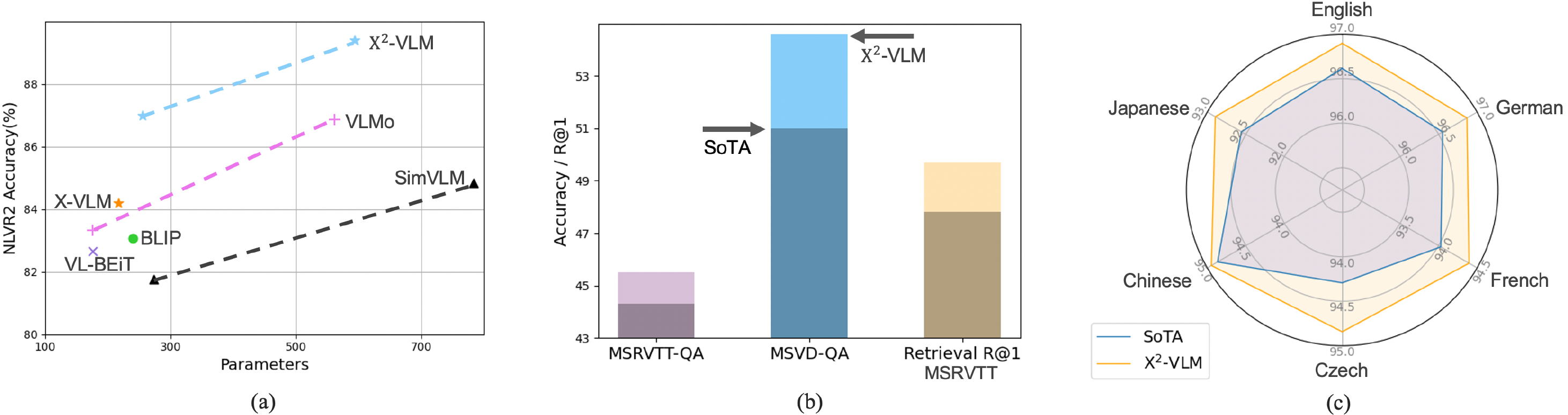}}
\caption{(a) Comparison of \baby with existing image-text pre-training methods on the visual reasoning task. (b) Comparison with existing video-text pre-training methods on video-text tasks. (c) Comparison with existing multilingual multi-modal pre-training methods.}
\vspace{-0.5cm}
\label{Fig:intro} 
\end{center}
\end{figure*}

Vision language pre-training aims to learn vision language alignments from a large number of image-text or video-text pairs. A pre-trained Vision Language Model (VLM) fine-tuned with a small amount of labeled data has shown state-of-the-art (SoTA) performances in many Vision Language (V+L) tasks, such as image-text retrieval and visual question answering (VQA).

Existing work learning vision language alignments typically falls into two categories: \textit{coarse-grained} and \textit{fine-grained}. Coarse-grained approaches use convolutional neural networks~\cite{he2016deep} or vision transformers~\cite{dosovitskiy2020image} to encode overall image features~\cite{huang2020pixel, kim2021vilt, li2021align}, which however have difficulties in learning fine-grained vision language alignments, e.g., at the object level, from noisy image-text pairs which are usually weak-correlated~\cite{huo2021wenlan}. To learn fine-grained vision language alignments, many approaches adopt pre-trained object detectors as the image encoder~\cite{tan2019lxmert, lu2019vilbert, li2019visualbert, gan2020large, chen2020uniter, li2020oscar, zhang2021vinvl}. However, object detectors output object-centric features unable to encode relations among multiple objects. Moreover, an object detector can only recognize a limited number of object categories.


Ideally, a VLM should simultaneously learn multi-grained alignments between vision and language in pre-training without being restricted to object-text alignments or image-text alignments. However, learning multi-grained alignments is challenging, and previous work has failed to handle this issue. The challenges come from four aspects: 1) what types of data to use to learn multi-grained vision language alignments; 2) how to aggregate the different types of data in a unified way for vision language pre-training; 3) how to represent multi-grained visual concepts, including objects, regions, and images, by a single vision encoder; 4) how to efficiently learn multi-grained vision language alignments from the data.


In this paper, we present an all-in-one VLM pre-trained with a unified framework to learn multi-grained vision language alignments, namely \baby. We leverage three types of data for vision language pre-training, including object labels on images~\cite{lin2014microsoft,shao2019objects365,kuznetsova2018open} such as ``man'' or ``backpack'', region annotations on images~\cite{krishna2016visual,kuznetsova2018open} such as ``boy wearing backpack'', and text descriptions for images such as ``The first day of school gives a mixed feeling to both students and parents.''. We assume that learning multi-grained vision language alignments can help VLMs better understand weak-correlated image-text pairs since the model has learned to align the components in images, e.g., objects or regions, to textual descriptions, e.g., words or phrases. We associate all visual concepts with text descriptions instead of class labels, including objects, regions, and images. By associating all visual concepts with language, the model can learn unlimited visual concepts described by diverse texts in a unified way.

\baby has a flexible modular architecture, with three modules for vision, text, and fusion, respectively. All modules are based on Transformer~\cite{vaswani2017attention}. We encode an image with vision transformer~\cite{dosovitskiy2020image}, and we utilize certain patch features to represent multi-grained visual concepts in the image that can be objects, regions, or the image itself. By doing so, \baby outputs vision features for objects, regions, and images in a unified form. Furthermore, we propose directly aligning the multi-grained vision features with the paired text features and simultaneously locating multi-grained visual concepts in the same image given different text descriptions for vision language pre-training. In fine-tuning and inference, \baby can leverage the learned multi-grained alignments to perform the downstream V+L tasks without object or region annotations in the input images.

\baby can be easily extended to video-text pre-training. For video encoding, we sample video frames and encode the frames with vision transformer respectively. Then, we use the average in the temporal dimension of patch features of frames to encode a video. The encoder parameters are shared between video-text pre-training and image-text pre-training. By doing so, we leverage video-text pairs to enable the model to understand visual concepts in temporal dimension and learn a more versatile VLM.

Moreover, we show the flexibility of \baby with the modular architecture. We investigate whether the cross-modal ability can be transferred to other languages or domains after pre-training. This is an important problem in real-world applications because many multi-modal tasks exist in non-English languages. However, since collecting image-text pairs or video-text pairs in certain languages or domains can be costly, recent SoTA VLMs are trained with English data and only applicable to English texts, limiting their application scopes. We find that surprisingly \baby can effectively adapt to V+L tasks in different languages or domains by simply replacing the text module with a language-specific or domain-specific one without further pre-training.


We conduct extensive experiments to verify the effectiveness of \baby. First, we compare \baby with SoTA image-text pre-training methods on base and large scale and find that \baby substantially outperforms all of them in the image-text tasks, including retrieval, VQA, reasoning, and grounding. Moreover, \baby outperforms SimVLM~\cite{wang2021simvlm} and BLIP~\cite{li2022blip}, which are designed for generative tasks, in image caption generation. \baby also outperforms MDETR~\cite{kamath2021mdetr} and OFA~\cite{wang2022ofa}, which also leverage image annotations of objects and regions, in cross-modal understanding tasks. \babyL with $\sim$590M parameters performs competitively to CoCa~\cite{yu2022coca} and BEiT-3~\cite{wang2022image} with $\sim$2B parameters, especially on image-text retrieval and visual reasoning. In summary, \baby makes a good trade-off between performance and model scale, as indicated in Figure \ref{Fig:intro} (a). Besides, we find that by training with large-scale image-text pairs, \baby learns to locate diverse fine-grained visual concepts in open-domain images, such as different sodas, cars, characters and celebrities. Second, \baby is also the new SoTA pre-trained model on video-text tasks, including video-text retrieval and video VQA, as shown in Figure \ref{Fig:intro} (b). Most existing VLMs only tackle image-text tasks, but \baby with a unified framework achieves SoTA performances on both types of tasks. Third, to verify the flexibility of the modular design, we replace the text encoder of \baby with XLM-R~\cite{conneau2020unsupervised}, a multilingual text encoder, after vision-language pre-training on English data. As indicated in Figure~\ref{Fig:intro} (c), \baby outperforms SoTA multilingual multi-modal pre-training methods that need multilingual image-text pairs~\cite{zhou2021uc2, jain2021mural} and multilingual sentence pairs~\cite{cclm} which are costly to collect.

The contributions of this paper are as follows:
\begin{itemize}

\item We propose to learn multi-grained vision language alignments by a unified pre-training framework that learns multi-grained aligning and multi-grained localization simultaneously. Based on it, we present \baby, an all-in-one pre-trained VLM that can handle both image-text and video-text tasks. 

\item Experiment results show that \baby is the best model on base and large scale on both image-text and video-text benchmarks. Furthermore, the results confirm that the proposed framework for multi-grained vision language pre-training is scalable to massive data and larger model sizes.

\item  We reveal the potential of the modular design of \baby, showing that it can be utilized in other languages or domains. By replacing the text encoder with XLM-R after pre-training on English data, \baby outperforms SoTA methods on multi-lingual multi-modal tasks.

\end{itemize}

\section{Related Work}
\label{sec:related}

\subsection{Image-Text Pre-training} 


The existing work on image-text pre-training typically falls into two categories: fine-grained and coarse-grained. Fine-grained approaches ~\cite{tan2019lxmert, lu2019vilbert, li2019visualbert, gan2020large, chen2020uniter, li2020oscar, zhang2021vinvl} utilize a pre-trained object detector~\cite{ren2015faster, anderson2018bottom} as the image encoder, which is trained on annotations of common objects, e.g. COCO~\cite{lin2014microsoft} and Visual Genome~\cite{krishna2016visual}. An object detector first identifies all regions that probably contain an object, then conducts object classification on each region. An image is then represented by dozens of object-centric features of the identified regions. However, object-centric features cannot represent relations among multiple objects in different regions. Therefore, it is difficult for this approach to effectively encode multi-grained visual concepts. Moreover, object detectors can only detect common objects, e.g. only 80 object categories for the COCO dataset. Thus, it is suboptimal to apply this approach to encode various visual concepts in real-world applications. For example, the approach cannot distinguish ``Pepsi'' from ``Coca Cola'' or ``Audi'' from ``BMW''. 

In contrast, the coarse-grained approaches build VLMs by extracting and encoding overall image features with convolutional network~\cite{jiang2020defense, huang2020pixel} or vision transformer~\cite{kim2021vilt, li2021align}. While being more efficient, the performance of the coarse-grained approach is usually not as good as the fine-grained approach since the latter leverages vision language alignments at the object level, which are shown to be critical for downstream V+L tasks. However, with advanced vision transformers, e.g. Swin-Transformer~\cite{liu2021swin} and BEiT-2~\cite{peng2022beit}, recent methods such as METER~\cite{dou2021empirical} and VL-BEiT~\cite{bao2022vl}, can outperform strongest fine-grained approach VinVL~\cite{zhang2021vinvl}.

There also emerge some methods attempting to learn both object-level and image-level alignments. However, these approaches still rely on object detectors and thus suffer from the aforementioned problems. For example, E2E-VLP~\cite{xu2021e2e} adds an end-to-end object detection module (i.e. DETR~\cite{carion2020end}). KD-VLP~\cite{liu2021kd} relies on external object detectors to perform object knowledge distillation. Different from these approaches, our framework for multi-grained vision language pre-training does not rely on object detection, and it learns vision language alignments not restricted to object-level or image-level in a unified way.

\subsection{Video-Text Pre-training}
Most existing VLMs only tackle image-text tasks. Only a few VLMs work on video-text pre-training. Since a video consists of multiple images, video-text models usually share many similarities with image-text models in both model architecture and training objectives. Though video-text pre-training shares similarities with image-text pre-training, no existing method can achieve SoTA performances on both types of tasks. Representative work on video-text pre-training including ClipBERT~\cite{lei2021less}, Frozen~\cite{bain2021frozen}, ALPRO~\cite{li2022align}, VIOLET~\cite{fu2021violet}, and All-in-one~\cite{wang2022all}. There are other methods optimized specifically for a downstream task, for either video-text retrieval~\cite{xue2022clip,min2022hunyuan_tvr} or video question answering~\cite{yang2022zero}. Recently, OmniVL~\cite{wang2022omnivl} is proposed to support both image-text tasks and video-text tasks. It utilizes 3D patch embeddings for videos and 2D patch embeddings for images, and adopts TimeSformer~\cite{bertasius2021space} for vision encoding.

\subsection{Multilingual Multi-modal Pre-training}
Multilingual multi-modal pre-training aims to make multi-modal models applicable to non-English texts. While appealing, multi-lingual multi-modal pre-training has its own challenges. Unlike multi-lingual pre-training and multi-modal pre-training where a relatively large amount of parallel data is available, there exist only a few multi-lingual multi-modal corpora and their language coverage is also limited. Therefore, $\text{M}^3\text{P}$~\cite{ni2021m3p} utilizes 101G texts covering 100 languages for pre-training. It makes English a pivot and alternates between English-only vision-language pre-training and multi-lingual masked language modeling. Differently, UC$^2$~\cite{zhou2021uc2} translates image-text pairs in English into five different languages and uses all the data for pre-training. MURAL~\cite{jain2021mural} collects large-scale image-text pairs in 110 languages. CCLM~\cite{cclm} utilizes parallel multilingual text pairs and proposes a simple framework that unifies cross-lingual and cross-modal pre-training with shared architecture and objectives. All these methods require extra data to perform multilingual multi-modal pre-training. In contrast, we show that \baby can adapt to multilingual V+L tasks without the need for multilingual multi-modal pre-training process by exploiting the potential of its modular architecture.
\begin{figure}[t]
\begin{center}
\centerline{\includegraphics[width=0.6\columnwidth]{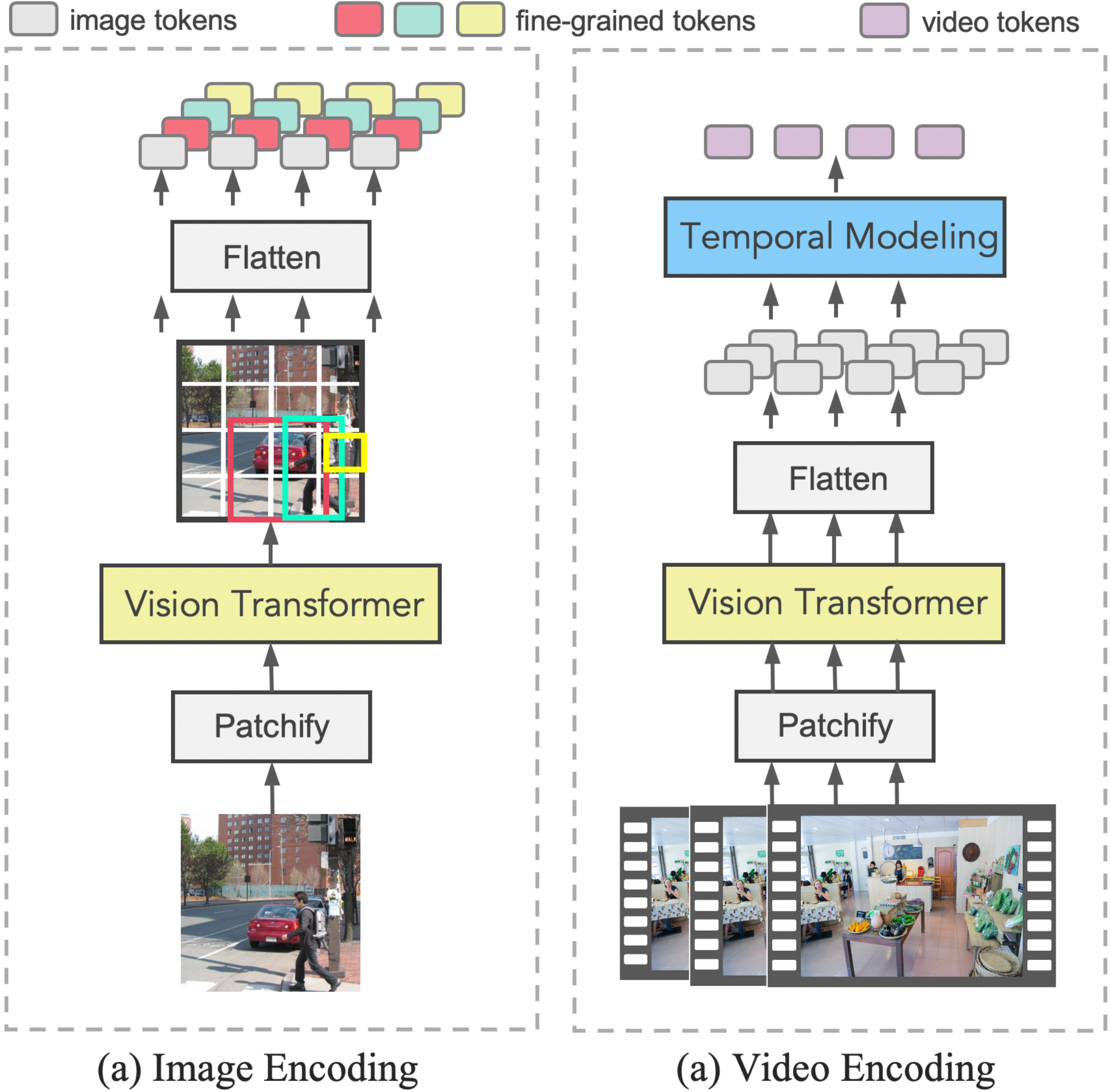}}
\caption{\textbf{Unified vision encoding} For images, we extract the subset of patch features from the vision transformer to represent an image and objects/regions in the image. For videos, each frame is first encoded independently, and than a light-weight non-parametric temporal modeling layer is applied across frames. }
\vspace{-0.5cm}
\label{Fig:vision}
\end{center}
\end{figure}

\section{Method}

\subsection{Overview}

\noindent\textbf{Architecture}: \baby consists of vision, text, and multi-modal fusion modules. All modules are based on Transformer~\cite{vaswani2017attention}. The fusion module takes text features as input and fuses the vision features with the text features through cross-attention at each layer, where the text features work as the queries and the vision features work as the keys and values. In pre-training, the three modules work as encoders, while the text and fusion modules can also be adapted for generation tasks if applying left-to-right self-attention as shown in our experiments for image caption generation. Figure~\ref{Fig:model} illustrates the architecture of \baby and the way we perform multi-grained aligning and multi-grained localization.

\noindent\textbf{Data}: \baby is a unified approach that associates all visual concepts with text descriptions, including image-text pairs, video-text pairs, and image annotations of objects and regions. That is to say, an image may contain more than one visual concept and each of them is associated with a text description, denoted as $(I, T, \{(V^{1}, T^{1}), ...\}^{N})$. $\{(V^{1}, T^{1}), (V^{2}, T^{2})...\}^{N}$ are the image annotations of objects or regions. Here, $V^{i}$ is an object or region in a bounding box $\mathbf{b}^{i}=(cx, cy, w, h)$ represented by the normalized center coordinates, width, and height of the box. When the image itself represents a visual concept, $\mathbf{b} = (0.5, 0.5, 1, 1)$. $T^{i}$ for objects are originally object labels. If an object annotation contains object attributes, e.g. color, we concatenate the attribute with the object label as the text description. $T^{i}$ for regions are phrases that describe the regions. Note that, as listed in Table~\ref{tbl:data}, some images do not have associated texts, i.e., $T$ is NaN, and some images do not have annotations, i.e., $N=0$. \textbf{Nevertheless, we mix all types of data in a training batch, and thus for each training iteration, we optimize the model by multi-grained aligning and multi-grained localization simultaneously. }


\subsection{Unified Vision Encoding}

\baby unifies image and video encoding, as illustrated in Figure~\ref{Fig:vision}. Irrespective of the inputs, the vision module of \baby produces hidden states in the latent feature space of the vision transformer. As a result, image-text pre-training and fine-grained pre-training mutually reinforce one another. Moreover, the capability of image understanding can be better transferred for video comprehension.


\noindent\textbf{Visual Concept Representation} \baby proposes an efficient way to obtain all multi-grained visual concepts in an image with only one forward pass of the vision transformer. First, we process an image into patch features. Then, \baby represents an object or a region, e.g. $V^{i}$, that corresponds to a set of patches in the bounding box, e.g. $\mathbf{b}^{i}$, by aggregating information among the patches as illustrated in Figure~\ref{Fig:vision}. Specifically, we flatten the corresponding patch features while keeping their original positions. Then, we calculate the average of the patch features as the [CLS] patch and prepend it. Accordingly, the representation of the entire image $I$ is obtained by aggregating information among all the patches. 

\noindent\textbf{Video Representation} 
Since a video consists of multiple images, to leverage large-scale image-text pre-training for better video understanding, we unify video encoding and image encoding in a simple and efficient way. First, we sample one frame per second for videos. Then, for each training iteration, we randomly sample a few frames of a video. The vision encoder will encode the frames into patch features respectively. Finally, we add temporal information to the patch features of each frame and calculate the average in temporal dimension to represent the video. By doing so, a video is encoded by a sequence of patch features the same as an object/region/image, and thus we can apply a unified pre-training framework for both video-text pairs and object/region/image-text pairs.

\subsection{Multi-Grained Vision Language Pre-training}
\label{sec:xvlm}

\begin{figure*}[t]
\begin{center}
\centerline{\includegraphics[width=1\columnwidth]{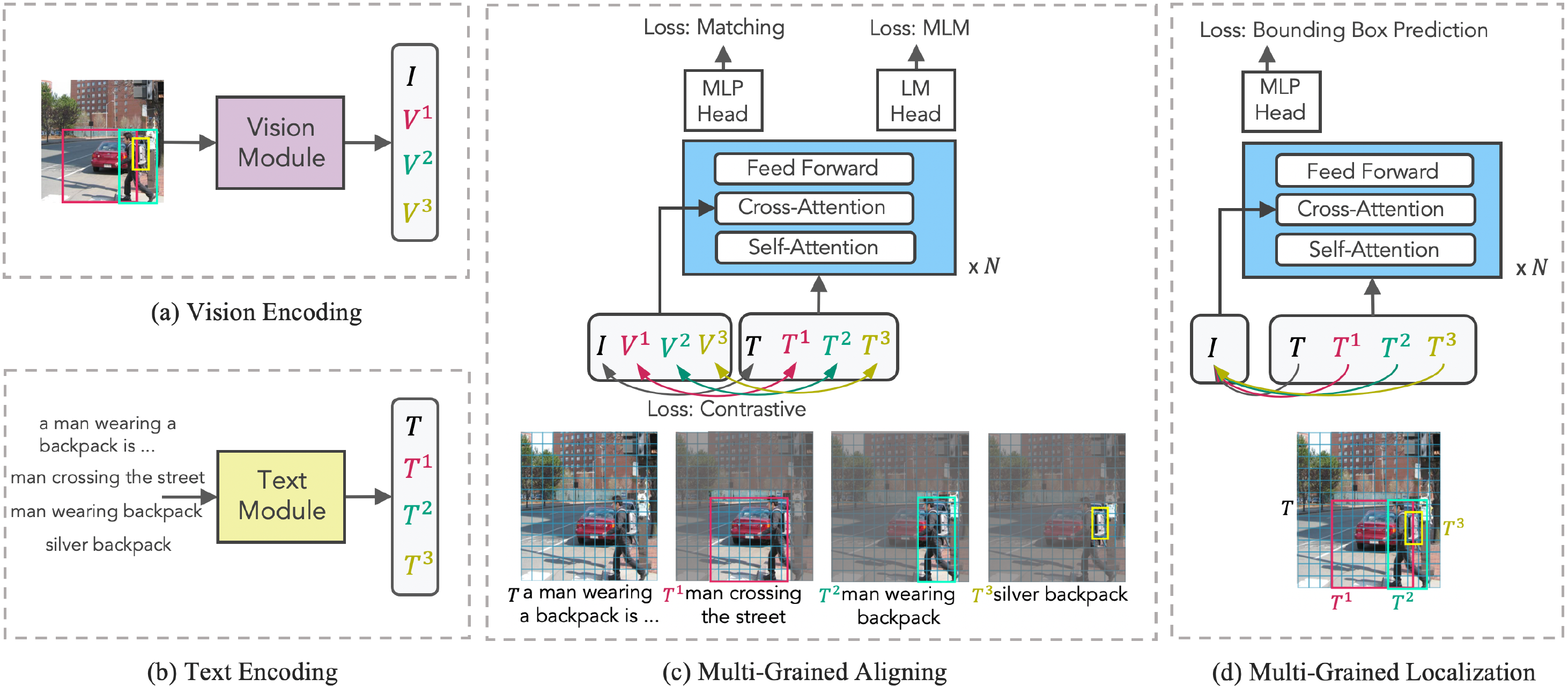}}
\caption{\textbf{Illustration of the proposed multi-grained vision language pre-training.} \baby consists of vision, text, and fusion modules. After encoding visual concepts (a) and text inputs (b), multi-grained vision features are then paired with corresponding text features for multi-grained aligning (c). Besides, the image is paired with different textual descriptions for multi-grained localization to predict the bounding box for each visual concept (d). All the datasets we used are publicly available (see Section~\ref{sec:pretraindata}).
}
\vspace{-0.5cm}
\label{Fig:model}
\end{center}
\end{figure*}

We mix all types of data in a training batch, and thus for each training iteration, as shown in Figure~\ref{Fig:model}, we optimize \baby by two objectives simultaneously: 1) learning multi-grained alignments between visual concepts and texts; 2) locating multi-grained visual concepts in images given different text descriptions.

\subsubsection{Multi-Grained Aligning}
Since we have associated all visual concepts with text descriptions, we propose to align the multi-grained visual concepts with the corresponding texts. Specifically, after encoding visual concepts by the aforementioned method, we align the vision features in multiple granularities with the corresponding text features in the same way. We simply choose three losses for optimization, including contrastive loss, matching loss, and MLM loss. These losses have been well-studied by previous work~\cite{chen2020uniter, radford2021learning, li2021align}, but we propose to employ them on the visual concept-to-text level. Note that $V$ in this section represents a visual concept, including an object, region, image, or video.

We apply contrastive loss to predict (visual concept, text) pairs from in-batch negatives. Given a pair $(V,T)$, $T$ is the positive example for $V$, and we treat the other $(N-1)$ texts within the mini-batch as negative examples. First, we define the similarity by: 
\begin{equation}
s(V,T) = g_v(\mathbf{v}_\mathrm{cls})^\top g_w(\mathbf{w}_\mathrm{cls}),
\end{equation}
where $\mathbf{v}_\mathrm{cls}$ and $\mathbf{w}_\mathrm{cls}$ are the output \texttt{[CLS]} embedding of the vision encoder and the text encoder respectively. $g_v$ and $g_w$ are transformations that map the \texttt{[CLS]} embeddings to normalized lower-dimensional representations. Based on it, we calculate the in-batch vision-to-text similarity as: 
\begin{equation}
p^\mathrm{v2t}(V) = \frac{\exp (s(V,T) / \tau)}{\sum_{i=1}^N \exp (s(V,T^i)/ \tau)},
\label{eq:pi2t}
\end{equation}
Similarly, the text-to-vision similarity is: 
\begin{equation}
p^\mathrm{t2v}(T) = \frac{\exp (s(V,T)/ \tau)}{\sum_{i=1}^N \exp (s(V^i,T)/ \tau)},
\end{equation}
where $\tau$ is a learnable temperature parameter. Let $\mathbf{y}^\mathrm{v2t}(V)$ and $\mathbf{y}^\mathrm{t2v}(T)$ denote the ground-truth one-hot similarity, in which only the positive pair has the probability of one. Finally, the contrastive loss is defined as the cross-entropy $\mathrm{H}$ between $\mathbf{p}$ and $\mathbf{y}$: 
\begin{equation}
\label{eqn:itc}
\mathcal{L}_\mathrm{cl} = \frac{1}{2} \mathbb{E}_{V,T\sim D} \big[ \mathrm{H}(\mathbf{y}^\mathrm{v2t}(V),\mathbf{p}^\mathrm{v2t}(V))  + \mathrm{H}(\mathbf{y}^\mathrm{t2v}(T),\mathbf{p}^\mathrm{t2v}(T)) \big]
\end{equation}


We also utilize the matching loss to determine whether a pair of visual concept and text is matched. For each visual concept in a mini-batch, we sample an in-batch hard negative text by following $p^\mathrm{v2t}(V)$ in Equation~\ref{eq:pi2t}. Texts that are more relevant to the concept are more likely to be sampled. We also sample one hard negative visual concept for each text. We then put the pairs as inputs for the fusion module, and then we use $\mathbf{x}_\mathrm{cls}$, the output \texttt{[CLS]} embedding of the fusion module, to predict the matching probability $p^\mathrm{match}$, and the loss is:
\begin{equation}
\label{eqn:itm}
\mathcal{L}_\mathrm{match} = \mathbb{E}_{V,T\sim D} \mathrm{H} (\mathbf{y}^\textrm{match}, \mathbf{p}^\textrm{match}(V,T)),
\end{equation}
where $\mathbf{y}^\textrm{match}$ is a 2-dimensional one-hot vector representing the ground-truth label.

Furthermore, we apply masked language modeling loss to predict the masked words in the text based on the visual concept. We randomly mask out the input tokens with a probability of 40\%, and the replacements are 10\% random tokens, 10\% unchanged, and 80\% \texttt{[MASK]}. We use the fusion encoder's outputs and append a linear layer followed by softmax for prediction. Let $\hat{T}$ denote a masked text, and $\mathbf{p}^j(V,\hat{T})$ denote the probability of the masked token $t_j$ predicted by the fusion module. We minimize the cross-entropy loss:
\begin{equation}
\label{eqn:mlm}
\mathcal{L}_\mathrm{mlm} = \mathbb{E}_{t_j \sim T; (V, T)\sim D} \mathrm{H} (\mathbf{y}^j, \mathbf{p}^j(V,\hat{T})),
\end{equation}
where $\mathbf{y}^j$ is a one-hot distribution in which the ground-truth token $t_j$ has the probability of one.

\subsubsection{Multi-Grained Localization}
We have aligned visual concepts with texts in different granularity. We further optimize \baby by training it to locate different visual concepts in the same image given corresponding text descriptions. Specifically, we introduce bounding box prediction task into vision language pre-training, where the model is asked to predict the bounding box $\mathbf{b}^{i}=(cx, cy, w, h)$ of a visual concept $V^i$: 
\begin{equation}
\hat{\mathbf{b}}^{i}(I, T^{i}) = \mathrm{Sigmoid}(\mathrm{MLP}(\mathbf{x}^{i}_\mathrm{cls})),
\end{equation}
where Sigmoid is for normalization, MLP denotes multi-layer perceptron, and $\mathbf{x}^{i}_\mathrm{cls}$ is the output \texttt{[CLS]} embedding of the fusion module given the features of $I$ (the entire image) and $T^{i}$ (the description of the visual concept).

For bounding box prediction, $\ell_1$ is the most commonly-used loss. However, it has different scales for small and large boxes, even if their relative errors are similar. To mitigate this issue, we use a linear combination of the $\ell_1$ loss and the generalized Intersection over Union (IoU) loss~\cite{rezatofighi2019generalized}, which is scale-invariant. The overall loss is defined as: 
\begin{equation}
\mathcal{L}_\mathrm{bbox} = \mathbb{E}_{(V^{i}, T^{i}) \sim I; I \sim D} [\mathcal{L}_\mathrm{iou}(\mathbf{b}^{i}, \hat{\mathbf{b}}^{i}) + ||\mathbf{b}^{i}- \hat{\mathbf{b}}^{i}||_1 ]
\end{equation}

Finally, the pre-training objective of \baby is defined as:
\begin{equation}
\mathcal{L} = \mathcal{L}_\mathrm{bbox} + \mathcal{L}_\mathrm{cl} + \mathcal{L}_\mathrm{match} + \mathcal{L}_\mathrm{mlm} 
\end{equation}

\section{Experiment}

\begin{table}[h]
\small
\centering	
\resizebox{0.7\columnwidth}{!}{%
\begin{tabular}	{ l | l |  l | l | l}
\toprule
Dataset & \# Images & \# Captions & \# Objects & \# Regions \\
\midrule
COCO & 0.11M & 0.55M & 0.45M & -\\
VG & 0.10M & - & 2.0M & 3.7M \\
SBU & 0.86M & 0.86M & - & -\\
CC-3M & 2.9M & 2.9M & - & - \\
\midrule
CC-12M & 11.1M & 11.1M & - & - \\
Objects365 & 0.58M  & - & 2.0M & - \\ 
OpenImages & 1.7M  & - & 4.2M & - \\
LAION & 1.3B & 1.3B & - & - \\
WebVid2.5M & 2.5M & 2.5M & - & - \\
Howto100M & 1.7M & 1.7M & - & - \\
YTT180M & 5.3M & 5.3M & - & - \\ 
 \bottomrule
\end{tabular}
}
\vspace{0.2cm}
\caption
{
\textbf{Statistics of the pre-training datasets.} We pre-train \baby with two sets of data: one contains COCO, VG, SBU, and CC-3M, where the total number of images is 4M; the other one includes more noisy image-text pairs and video-text pairs.  
}  
\label{tbl:data}
\end{table}

\begin{table}[t]
\centering
\resizebox{0.7\columnwidth}{!}{%
\begin{tabular}{@{\hskip1pt}l@{\hskip1pt} @{\hskip1pt}c@{\hskip2pt} @{\hskip2pt}c@{ \hskip1pt} @{\hskip1pt}c@{ \hskip1pt} @{\hskip2pt}c@{ \hskip1pt} @{\hskip1pt}c@{ \hskip1pt} @{\hskip2pt}c@{ \hskip1pt} @{\hskip1pt}c@{ \hskip1pt} }
\toprule
\multirow{3}{*}{Model} & \multirow{3}{*}{Hidden} & \multicolumn{2}{c}{Vision} & \multicolumn{2}{c}{Text} & \multicolumn{2}{c}{Fusion} \\ 
\cmidrule(lr){3-4} \cmidrule(lr){5-6} \cmidrule(lr){7-8}
 &  & Layers & Params & Layers & Params & Layers & Params \\ 
\midrule
\babyB & 768 & 12 & 86M & 12 & 111M & 6 & 55M \\
\babyL & 1024 & 24 & 303M & 12 & 190M & 6 & 95M \\ 
\bottomrule
\end{tabular}
}
\vspace{0.2cm}
\caption{\textbf{Size variants of \babyx.} All modules consist of transformer layers.}
\label{tab:modelsize}
\end{table}

\begin{table*}[t]
\centering
\resizebox{1\columnwidth}{!}{%
\begin{tabular}{@{\hskip1pt}l@{\hskip1pt} @{\hskip1pt}c@{\hskip1pt} @{\hskip1pt}c@{ \hskip1pt} @{\hskip1pt}c@{ \hskip1pt} @{\hskip1pt}c@{ \hskip1pt} @{\hskip1pt}c@{ \hskip1pt} @{\hskip1pt}c@{ \hskip1pt} @{\hskip1pt}c@{ \hskip1pt} | @{ \hskip2pt}c@{ \hskip1pt} @{\hskip1pt}c@{ \hskip1pt} @{\hskip1pt}c@{ \hskip1pt} @{\hskip1pt}c@{ \hskip1pt} @{\hskip1pt}c@{ \hskip1pt} @{\hskip1pt}c@{ \hskip1pt} }
\toprule
\multirow{3}{*}{  Model} & \multirow{3}{*}{ \# Params} & \multicolumn{6}{c}{ MSCOCO (5K test set)} & \multicolumn{6}{c}{ Flickr30K (1K test set)} \\
 & & \multicolumn{3}{c}{ TR} & \multicolumn{3}{c}{  IR} & \multicolumn{3}{c}{ TR} & \multicolumn{3}{c}{ IR} \\
 & & R@1 & R@5 & R@10 & R@1 & R@5 & R@10 & R@1 & R@5 & R@10 & R@1 & R@5 & R@10 \\
\midrule
\multicolumn{14}{l}{\textit{Models Pretrained on COCO, VG, SBU and CC datasets (4M)}} \\
ALBEF & 210M & 73.1 & 91.4 & 96.0 & 56.8 & 81.5 & 89.2 & 94.3 & 99.4 & 99.8 &82.8 & 96.7 & 98.4 \\
VLMo$_\mathrm{base}$$\dag$ & 175M &  74.8 & 93.1 & 96.9 &  57.2 &  82.6 & 89.8 & 92.3 & 99.4 & 99.9 & 79.3 & 95.7 & 97.8 \\
VL-BEiT & 175M & 79.5 & - & - & 61.5 & - & - & 95.8 & - & - & 83.9 & - & - \\
OmniVL & 288M & 76.8 & 93.6 & 97.3 & 58.5 & 82.6 & 89.5 & 94.9 & 99.6 & 99.9 & 83.4 & 97.0 & 98.6 \\

\bf \babyx$_\mathrm{base}$ & 255M & \bf 80.5 & \bf 95.5 & \bf 97.8 & \bf 62.7 & \bf 84.7 & \bf 90.7 & \bf 97.4 & \bf 99.9 & \bf 100 & \bf 90.0 & \bf 98.6 & \bf 99.3 \\

\midrule
VLMo$_\mathrm{large}$$\dag$ & 562M & 78.2 & 94.4 & 97.4 & 60.6 & 84.4 & 91.0 & 95.3 & 99.9 & 100 &  84.5 &  97.3 &  98.6 \\

\bf \babyx$_\mathrm{large}$ & 593M & \bf 82.3 & \bf 96.2 & \bf 98.3 & \bf 65.2 & \bf 86.4 & \bf 91.9 & \bf 99.1 & \bf 100 & \bf 100 & \bf 91.1 & \bf 98.6 & \bf 99.4 \\
\midrule
\midrule
\multicolumn{14}{l}{\textit{ Models Pretrained on More Data}} \rule{0pt}{2.5ex} \\

BLIP$_\mathrm{base}$ & 240M  & 81.9 & 95.4 & 97.8 & 64.3 & 85.7 & 91.5 & 97.3 & 99.9 & 100 & 87.3 & 97.6 & 98.9 \\
OmniVL & 288M & 82.1 & 95.9 & 98.1 & 64.8 & 86.1 & 91.6 & 97.3 & 99.9 & 100 & 87.9 & 97.8 & 99.1 \\

\bf \babyx$_\mathrm{base}$ & 255M & \bf 83.5 & \bf 96.3 & \bf 98.5 & \bf 66.2 & \bf 87.1 & \bf 92.2 & \bf 98.5 & \bf 100 & \bf 100 & \bf 90.4 & \bf 98.2 & \bf 99.3 \\

\midrule
ALIGN$\dag$ & 490M & 77.0 & 93.5 & 96.9 & 59.9 & 83.3 & 89.8 & 95.3 & 99.8 & 100 & 84.9 & 97.4 & 98.6 \\
FLIP$\dag$ & 420M & 78.9 & 94.4 &  97.4 & 61.2 & 84.3 & 90.6 & 96.6 & 100 & 100 &  87.1 & 97.7 & 99.1 \\

BLIP$_\mathrm{large}$ & 452M & 82.4 & 95.4 & 97.9 & 65.1 & 86.3 & 91.8 & 97.4 & 99.8 & 99.9 & 87.6 & 97.7 & 99.0 \\

\bf \babyx$_\mathrm{large}$ & 593M & \bf 84.4 & \bf 96.5 & \bf 98.5 & \bf 67.7 & \bf 87.5 & \bf 92.5 & \bf 98.8 & \bf 100 & \bf 100 & \bf 91.8 & \bf 98.6 & \bf 99.5 \\


\bottomrule
\end{tabular}
}
\caption{\textbf{Results of image-to-text retrieval (TR) and
text-to-image retrieval (IR) on COCO and Flickr30K.} $\dag$ denotes dual-encoder retrieval models, and others use a fusion module to re-rank top-k candidates following ALBEF~\cite{li2021align}. 
}
\label{tbl:retrieval}
\end{table*}

\subsection{Pre-training Datasets}
\label{sec:pretraindata}

We pre-train \baby with two sets of data. The 4M pre-training dataset consists of two in-domain datasets, COCO~\cite{lin2014microsoft} and Visual Genome (VG)~\cite{krishna2016visual}, and two out-of-domain datasets, SBU Captions~\cite{ordonez2011im2text} and Conceptual Captions (CC)~\cite{sharma2018conceptual}. This pre-training dataset is widely utilized by previous work, and thus we choose this setting to make a fair comparison with other methods. We also include annotations for COCO and VG images from RefCOCO~\cite{yu2016modeling}, GQA~\cite{hudson2019gqa}, and Flickr entities~\cite{plummer2015flickr30k} following OFA~\cite{wang2022ofa} and MDETR~\cite{kamath2021mdetr}.

Then, we scale up the pre-training dataset by including out-of-domain and much noisier image-text pairs from Conceptual 12M dataset (CC-12M)~\cite{changpinyo2021conceptual} and LAION~\cite{schuhmann2022laion}, and object annotations from Objects365~\cite{shao2019objects365} and OpenImages~\cite{kuznetsova2018open}. Besides, to support video-text downstream tasks, we include video-text pairs from WebVid2.5M~\cite{bain2021frozen}, Howto100M~\cite{miech2019howto100m}, and YT-Temporal 180M~\cite{zellers2021merlot} for pre-training. Note that all the datasets we used are public available and have been exploited in previous work~\cite{li2021align,zhang2021vinvl,li2022blip,wang2022ofa,wang2022all}. Besides, since most downstream tasks are built on top of COCO and VG, we exclude all images that also appear in the test sets of downstream tasks to avoid information leak. We give data filtering details in Appendix.

\subsection{Implementation Details}
\label{sec:details}

Table~\ref{tab:modelsize} lists the parameters of \baby. Considering the trade-off between performance and model scale~\cite{wang2022efficientvlm}, \babyL also uses a 12L text encoder. The vision encoder is initialized with BEiT-2~\cite{peng2022beit}. The text encoder is initialized with BERT~\cite{devlin2019bert}. \baby is pre-trained at image resolution of $224\times224$ using $16\times16$ patch size. We mix all types of data in a training batch, and thus for each training iteration, we optimize the model by multi-grained aligning and multi-grained localization simultaneously. With 4M data, we pre-train \babyB for 500K steps with a batch size of 1024 on 8 A100 and \babyL for 250K steps on 16 A100, which takes $\sim1$ week. The learning rate of \babyB is warmed-up to $1e^{-4}$ in the first 2500 steps and decayed following a linear schedule. The learning rate is $5e^{-5}$ for \babyL. With large-scale data, training \baby takes 2-3 weeks on 32 A100 for the base model and 64 A100 for the large model. We describe the implementation details in Appendix.

\begin{table}[t]
\centering
\resizebox{0.6\columnwidth}{!}{%
\begin{tabular}{lccccc}
\toprule
\multirow{2}{*}{ } & \multirow{2}{*}{\# Params} & \multicolumn{2}{c}{ MSCOCO} & \multicolumn{2}{c}{ Flickr30K} \\
 & & TR &  IR &  TR & IR \\
\midrule
BEiT-3 & 1.9B & \bf 84.8 & 67.2 & 98.0 & 90.3 \\
\bf \babyx$_\mathrm{large}$ & 593M & 84.4 & \bf 67.7 & \bf 98.8 & \bf 91.8 \\
\bottomrule
\end{tabular}
}
\vspace{0.2cm}
\caption{\baby compared with the SoTA giant model, BEiT-3, on image-text retrieval benchmarks. We report Recall@1 for both image-to-text retrieval (TR) and text-to-image retrieval (IR).
}
\label{tbl:beit3}
\end{table}

\begin{table*}[!t]

	\centering	
\resizebox{1\columnwidth}{!}{%
	\begin{tabular}	{@{\hskip1pt}l@{\hskip1pt} @{\hskip1pt}c@{\hskip1pt} @{\hskip5pt}c@{ \hskip1pt} @{\hskip1pt}c@{ \hskip1pt} @{\hskip5pt}c@{ \hskip1pt} @{\hskip1pt}c@{ \hskip1pt} @{\hskip5pt}c@{ \hskip1pt} @{\hskip1pt}c@{ \hskip1pt} @{\hskip1pt}c@{ \hskip1pt} @{\hskip5pt}c@{ \hskip1pt} @{\hskip1pt}c@{ \hskip1pt}} 
	\toprule
	 \multirow{2}{*}{Method} & \multirow{2}{*}{\# Params} & \multicolumn{2}{c}{VQA} & \multicolumn{2}{c}{NLVR2} & \multicolumn{3}{c}{RefCOCO+} & \multicolumn{2}{c}{COCO Caption} \\
	  & & test-dev & test-std & dev & test-P & val & testA$^d$ & testB$^d$ & BLEU@4 & CIDEr \\
\midrule
\multicolumn{10}{l}{\textit{Models Pretrained on COCO, VG, SBU and CC datasets (4M)}} \\
ALBEF & 210M & 74.5 & 74.7 & 80.2 & 80.5 & - & - & - & - & - \\
VLMo$_\mathrm{base}$ & 175M & 76.6 & 76.9 & 82.8 & 83.3 & - & - & - & - & - \\
METER & 341M & 77.7 & 77.6 & 82.3 & 83.1 & - & - & - & - & - \\
VL-BEiT & 175M & 77.5 & 77.8 & 81.9 & 82.7 & - & - & - & - & - \\

\bf \babyx$_\mathrm{base}$ & 255M & \bf 79.2 & \bf 79.3 & \bf 85.9 & \bf 86.1 & \bf 85.4 & \bf 89.2 & \bf 77.3 & \bf 41.0 & \bf 133.6 \\
\midrule

VLMo$_\mathrm{large}$ & 562M & 79.9 & 80.0 & 85.6 & 86.9 & - & - & - & - & -  \\

\bf \babyx$_\mathrm{large}$ & 593M & \bf 80.5 & \bf 80.5 & \bf 87.2 & \bf 87.6 & \bf 86.9 & \bf 90.1 & \bf 80.2 & \bf 42.0 & \bf 136.7 \\

\midrule
\midrule

\multicolumn{10}{l}{\textit{Models Pretrained on More Data}} \\
OmniVL & 288M & 78.3 & 78.4 & - & - & - & - & - & 39.8 & 133.9 \\
SimVLM$_\mathrm{base}$ & 273M  & 77.9 & 78.1 & 81.7 & 81.8 & - & - & - & 39.0 & 134.8 \\
OFA$_\mathrm{base}$ & 182M & 78.0 & 78.1 & - & - & 81.4 & 87.2 & 74.3 & 41.0 & \bf 138.2 \\
BLIP$_\mathrm{base}$ & 240M & 78.2 & 78.2 & 82.5 & 83.1 & - & - & - & 39.4 & 131.4 \\

\bf \babyx$_\mathrm{base}$ & 255M & \bf 80.4 & \bf 80.2 & \bf 86.2 & \bf 87.0 & \bf 85.2 & \bf 90.3 & \bf 78.4 & \bf 41.7 & 136.1 \\

\midrule

SimVLM$_\mathrm{large}$ & 783M & 79.3 & 79.6 & 84.1 & 84.8 & - & - & - & 40.3 & \bf 142.6 \\

OFA$_\mathrm{large}$ & 472M & 80.3 & 80.5 & - & - & 85.8 & 89.9 & 79.2 & 42.4 & 142.2 \\

\bf \babyx$_\mathrm{large}$ & 593M & \bf 81.9 & \bf 81.8 & \bf 88.7 & \bf 89.4 & \bf 87.6 & \bf 92.1 & \bf 81.8 & \bf 42.6 & 139.1 \\


\bottomrule  	  
\end{tabular}
}
\caption
{
\textbf{Results on downstream image-text tasks}, including visual question answering (VQA), visual reasoning (NLVR2), visual grounding (RefCOCO+), and image caption generation (COCO Caption). 
}
\vspace{-0.3cm}
\label{tbl:results}
\end{table*}

\subsection{Image-Text Downstream Tasks}
We compare \baby with the most well-known state-of-the-art approaches on five widely used image-text downstream tasks. 
In general, we follow the settings in the previous work on fine-tuning. We describe how we implement fine-tuning as follows. 

\subsubsection{Image-Text Retrieval}

We evaluate \baby on both MSCOCO and Flickr30K~\cite{plummer2015flickr30k} datasets. We adopt the widely used Karpathy split~\cite{karpathy2015deep} for both datasets. We optimize $\mathcal{L}_\mathrm{cl}$ and $\mathcal{L}_\mathrm{match}$ for fine-tuning. We set the batch size to 1024. The resolution of input images is set to 384x384. Following the previous work~\cite{li2021align}, \baby first encodes images and texts separately and calculates in-batch text-to-image and image-to-text similarities to obtain the top-$k$ candidates, and then uses the fusion encoder to re-rank the candidates. $k$ is set to 80 for the MSCOCO dataset and 32 for Flickr30K.

Table~\ref{tbl:retrieval} shows that \baby achieves SoTA results on image-text retrieval tasks especially on Flickr30K benchmark even though existing approaches either have more model parameters or more training data. Concretely, \babyB outperforms FLIP~\cite{yao2021filip}, BLIP$_\mathrm{base}$ and BLIP$_\mathrm{large}$ which also exploits large-scale image-text pairs from LAION, and \babyL further improves the image-text retrieval performance. Compared to OmniVL which also supports both image-text tasks and video-text tasks, \babyB substantially outperforms it when pre-trained with the 4M data or with more data. These results validate the advantage of learning multi-grained vision language alignments.

We also compare \babyL with BEiT-3, a giant foundation model with 1.9B model parameters in Table~\ref{tbl:beit3}. Experimental results show that though being much smaller, \babyL has a comparable or even better performance compared with BEiT-3. Moreover, as shown in Table~\ref{tbl:retrieval}, \babyB substantially outperforms VL-BEiT which is the base version of BEiT-3 in the 4M setting. On the other hand, when comparing \baby's performances in different settings in Table~\ref{tbl:retrieval}, we can see that the proposed framework for multi-grained vision language pre-training has good scalability which can benefit from a larger model size and large-scale out-of-domain image-text pairs.

\subsubsection{Visual Question Answering}
The task requires the model to predict an answer given an image and a question. We evaluate \baby on the VQA v2.0 dataset~\cite{goyal2017making}. Following existing methods~\cite{tan2019lxmert, chen2020uniter, li2021align}, we use both train and validation sets for training and include additional question-answer pairs from Visual Genome. Following ALBEF, we use a six-layer Transformer decoder to generate answers based on the outputs of the fusion module. Then, the model is fine-tuned by optimizing the auto-regressive loss. During inference, we constrain the decoder to only generate from the 3,129 candidate answers to make a fair comparison with existing methods. Note that there is a NULL answer. Thus, the actual number of candidate answers is 3,128. Following previous work~\cite{wang2022ofa,yu2022coca,wang2022image}, the resolution of input images is set to 768x768.

We report the experimental results of VQA in Table~\ref{tbl:results}. We can see that \babyB and \babyL outperforms other approaches with similar scale of model size. Specifically, \babyB substantially outperforms ALBEF, VLMo, METER, and VL-BEiT in the 4M setting. Besides, with more pre-training data, \babyB outperforms BLIP which also exploits large-scale image-text pairs from LAION. Compared to OmniVL which also supports both image-text tasks and video-text tasks, \babyB substantially outperforms it, achieving an absolute improvement of 2\%. \baby also substantially outperforms SimVLM and OFA on both base and large scales. SimVLM utilizes an in-house 1.8B image-text dataset. OFA also leverages image annotations of objects and regions the same as \baby. These results confirm the effectiveness of the proposed framework for multi-grained vision language pre-training. Furthermore, when comparing \baby's performances in different settings in Table~\ref{tbl:results}, we can see that the proposed framework has good scalability which can benefit from a larger model size. When pre-training a larger model with more data, the performance improvement is even more remarkable.

\subsubsection{Visual Reasoning}
We evaluate \baby on widely used benchmark NLVR2~\cite{suhr2018corpus}. The task lets the model determine whether a text describes the relations between two images. Following previous work~\cite{wang2021vlmo, bao2022vl}, we formulate the triplet input to two image-text pairs, each containing the text description and one image. We then concatenate the final output [CLS] features of the fusion module of the two pairs to predict the label. The resolution of input images is set to 384x384. Given the results in Table~\ref{tbl:results}, we can observe that the visual reasoning task benefits more from the model size than the pre-training data scale. Comparing to other base-scale models, e.g. ALBEF, VLMo, VL-BEiT, SimVLM, and BLIP, \babyB has much better performance, achieving $\sim$ 3-4\% absolute improvement, no matter when pre-training with 4M data or with much more noisy data. \babyL also substantially outperforms other large-scale models, including VLMo$_\mathrm{large}$ and SimVLM$_\mathrm{large}$.

\subsubsection{Visual Grounding}
We evaluate \baby on RefCOCO+~\cite{yu2016modeling}. Given an image as the input and a text description as the query, the final output [CLS] features of the fusion module is utilized to predict the bounding box of the visual concept. The resolution of input images is set to 384x384. As indicated in Table~\ref{tbl:results}, \baby outperforms OFA~\cite{wang2022ofa} which also utilizes image annotations of objects and regions for pre-training. Differently, OFA with an encoder-decoder architecture formulates all the data in the form of sequence-to-sequence. 
Furthermore, \baby for general V+L purposes outperforms MDETR~\cite{kamath2021mdetr} specialized for visual grounding tasks, achieving absolute improvements of $\sim 7\%$ (average on metrics). These results confirm the effectiveness of the proposed multi-grained vision language pre-training compared to other approaches that also leverage image annotations of objects and regions.

\subsubsection{Image Captioning}
The task requires a model to generate text descriptions of input images. Though \baby is more for cross-modal understanding, we also evaluate its generation performance on the COCO Captioning dataset~\cite{chen2015microsoft}. Following UniLM~\cite{dong2019unified} and BEiT-3, we use left-to-right MLM for generation. Specifically, we employ the text module and fusion module as decoder with left-to-right self-attention and adopt the method~\cite{zeng-nie-2021-investigation} that decreases finetune-generation discrepancy in MLM generation. The resolution of input images is set to 480x480. We report BLEU-4 and CIDEr scores on the Karparthy test split. As shown in Table~\ref{tbl:results}, \babyB outperforms BLIP~\cite{li2022blip} and SimVLM~\cite{wang2021simvlm} which are designed for generative tasks. BLIP exploits large-scale image-text pairs from LAION the same as \baby. SimVLM utilizes an in-house 1.8B image-text dataset. \baby also outperforms OFA~\cite{wang2022ofa} in image captioning in terms of BLEU-4. OFA has an encoder-decoder architecture and formulates all downstream tasks into sequence-to-sequence form for pre-training. 
In general, though \baby is more for cross-modal understanding, it performs competitively or sometimes better compared with SoTA generative methods.

\begin{table}[t]
\centering
\resizebox{0.8\columnwidth}{!}{%
\begin{tabular}{lccccccc}
\toprule
\multirow{2}{*}{ } & \multicolumn{3}{c}{\textbf{Winoground}} & \multicolumn{4}{c}{\textbf{OVAD}}\\
 & Group & Text & Image & All & Head & Medium & Tail \\
\midrule

Random & 12.5 & 25.0 & 25.0 & 8.6 & 36.0 & 7.3 & 0.6 \\

\midrule
CLIP & 8.0 & 30.7 & 10.5 & 17.0 & 44.3 & 18.4 & 5.5 \\
ALBEF$_\text{4M}$ & 11.0 & 29.2 & 15.5 & 15.6 & 43.1 & 17.3 & 3.7 \\
BLIP & 11.7 & 35.5 & 15.0 & 24.3 & 51.0 & 28.5 & 9.7 \\
BLIP-2 & 18.2 & 43.0 & 22.0 & - & - & - & - \\
UNITER$_\mathrm{large}$ & 10.5 & 38.0 & 14.0 & - & - & - & - \\
PEVL & 12.2 & 33.2 & 15.7 & - & - & - & - \\
OVADetector & - & - & - & 21.4 & 48.0 & 26.9 & 5.2 \\ 
\midrule
\textbf{\babyx$_\mathrm{base}$}$_\text{4M}$ & 22.5 & 46.3 & 25.3 & 24.0 & 51.9 & 29.7 & 7.2 \\
\textbf{\babyx$_\mathrm{large}$}$_\text{4M}$ & 25.5 & 49.5 & 31.0 & 27.7 & 54.0 & 34.4 & 10.1 \\
\bf \babyx$_\mathrm{base}$ & 24.5 & 47.3 & 29.8 & 27.6 & 52.2 & 34.7 & 10.3 \\
\bf \babyx$_\mathrm{large}$ & \bf 25.8 & \bf 52.5 & \bf 32.5 & \bf 29.2 & \bf 55.1 & \bf 36.4 & \bf 11.3 \\

\bottomrule
\end{tabular}
}
\vspace{0.2cm}
\caption{\textbf{Zero-shot evaluation results on fine-grained downstream tasks}: Winoground, a fine-grained image-text matching task, and OVAD, Open-vocabulary Attribute Detection(mAP). 
}
\label{tbl:finegrained}
\end{table}

\subsubsection{Winoground}
Winoground~\cite{thrush2022winoground} presents a challenging task: given two images and two captions, the goal is to match them correctly, where the captions contain identical sets of words, but in a different order. Three metrics, namely Text (whether a model can match the correct caption for a given image), Image (vice versa), and Group (whether a model can match each pair), are used to evaluate the performance. Several competitive VLMs have been shown to perform close to or even below random chance. Experimental results in Table~\ref{tbl:finegrained} shows that even when trained on 4M data \baby substantially outperforms other models such as UNITER$_\mathrm{large}$, which is based on a large pre-trained object detector, and BLIP-2, which consists of giant ViT and FlanT5 large language model~\cite{chung2022scaling} and is pre-trained on a much larger dataset with 129M images. Notably, the performance of \baby can be further improved by increasing the model size or pre-training dataset.

\subsubsection{Open-vocabulary Attribute Detection}
Open-Vocabulary Attribute Detection (OVAD)~\cite{bravo2023open} aims to recognize an open set of objects in an image together with an open set of attributes for every object. We follow the benchmark and evaluate zero-shot performance of vision language models on attributes in the box-oracle setting. The experimental results are given in Table~\ref{tbl:finegrained}. \babyB pre-trained with 4M dataset has already been comparable to BLIP pre-trained with 129M dataset. \babyB also outperforms OVADetector which consists of a frozen CLIP text encoder and an object detector based on Faster-RCNN. Moreover, scaling \baby with larger pre-training datasets or larger model size consistently improve its performance as in other tasks.

\begin{table}[t]
\centering
\resizebox{0.7\columnwidth}{!}{%
\begin{tabular}{@{\hskip2pt}l@{\hskip2pt} @{\hskip2pt}c@{\hskip2pt} @{\hskip5pt}c@{\hskip2pt} @{\hskip2pt}c@{\hskip5pt} @{\hskip2pt}c@{ \hskip2pt} @{\hskip2pt}c@{\hskip2pt} @{\hskip2pt}c@{\hskip2pt} }
\toprule
\multirow{2}{*}{\bf Model} & \multirow{2}{*}{\textbf{\# Params}} & \multicolumn{2}{c}{\bf Video-QA} & \multicolumn{3}{c}{\bf MSRVTT (1K test set)} \\
 & & MSRVTT & MSVD & R@1 & R@5 & R@10 \\
\midrule
ALPRO & 513M & 42.1 & 45.9 & - & - & - \\  
VIOLET & 163M & 43.9 & 47.9 & - & - & - \\  
All-in-one & 110M & 44.3 & 47.9 & 37.9 & 68.1 & 77.1  \\
OmniVL & 288M & 44.1 & 51.0 & 47.8 & 74.2 & 83.8  \\
\midrule
\bf \babyB & 255M & 45.0 & 52.8 & 47.6 & 74.1 & 84.2 \\
\bf \babyL & 593M & \bf 45.5 & \bf 54.6 & \bf 49.6 & \bf 76.7 & \bf 84.2 \\

\bottomrule
\end{tabular}
}
\vspace{0.2cm}
\caption{\textbf{Fine-tuning results on video-text tasks}, including video question answering on MSRVTT and MSVD datasets, and text-to-video retrieval on MSRVTT. We report classification accuracy for VQA and Recall@K for text-to-video retrieval. }
\label{tbl:video}
\end{table}

\subsection{Video-Text Downstream Tasks}

\baby unifies image-text and video-text pretraining. In this section, we evaluate \baby on three widely used video-text tasks, including both \textbf{Video-Text Retrieval} (MSRVTT~\cite{xu2016msr}) and \textbf{Video Question Answering} (MSRVTT-QA~\cite{xu2017video} and MSVD-QA~\cite{xu2017video}). We implement a text-to-video retrieval model the same as image-text retrieval by first calculating top-$k$ candidates and then re-ranking the candidates using the fusion module. $k$ is set to 32. During training and inference, we sample five frames for each video. The image resolution is set to 384. Video question answering requires a model to generate an answer given a video and a question. Following previous work, we formulate it as a classification task given candidate answers. During training and inference, we sample five frames for each video in the MSRVTT dataset, and eight frames for the MSVD dataset. The image resolution is set to 320 for MSRVTT and 224 for MSVD. We compare with SoTA video-language foundation models: ALPRO~\cite{li2022align}, VIOLET~\cite{fu2021violet}, and All-in-one~\cite{wang2022all}. We also compare \baby with OmniVL which also supports both image-text tasks and video-text tasks. There are other methods optimized specifically for either video-text retrieval~\cite{xue2022clip,min2022hunyuan_tvr} or video question answering~\cite{yang2022zero}, which are not included in our comparison.

The results are given in Table~\ref{tbl:video}. We can see that \babyB outperforms previous video-language foundation models on both video question answering and text-to-video retrieval, and \babyL further advance the performance, achieving new SoTA results of video-text pre-training. Besides, we compare \baby with OmniVL on both image-text (Table~\ref{tbl:retrieval} and Table~\ref{tbl:results}) and video-text benchmarks. In general, \babyB substantially outperforms OmniVL on all image-text downstream tasks, including image-text retrieval, visual question answering, image caption generation, and video question answering.

\subsection{Multilingual Multi-modal Tasks}

\begin{table}[t]
\centering
\resizebox{0.7\columnwidth}{!}{%
\begin{tabular}{@{\hskip1pt}l@{\hskip5pt} @{\hskip5pt}c@{\hskip5pt} @{\hskip5pt}c@{ \hskip5pt} @{\hskip5pt}c@{ \hskip5pt} @{\hskip5pt}c@{ \hskip5pt} @{\hskip5pt}c@{ \hskip5pt} @{\hskip5pt}c@{ \hskip5pt} @{\hskip5pt}c@{ \hskip5pt}}

\toprule
\multirow{2}{*}{\textbf{Model}} & \multicolumn{4}{c}{ \textbf{Flickr30K}} & \multicolumn{3}{c}{\textbf{MSCOCO}} \\
  & EN  & DE & FR & CS & EN & ZH & JA  \\ 
\midrule

\multicolumn{8}{l}{\textit{Models with Multilingual Multimodal Pretraining}} \\

$\text{M}^3\text{P}$ & 87.7 & 82.7 & 73.9 & 72.2 & 88.7 & 86.2 & 87.9  \\
UC$^2$ & 88.2 & 84.5 & 83.9 & 81.2 & 88.1 & 89.8 & 87.5  \\ 
MURAL$_\mathrm{base}$$\dag$ & 92.2 & 88.6 & 87.6 & 84.2 & 88.6 & - & 88.4 \\
MURAL$_\mathrm{large}$$\dag$ & 93.8 & 90.4 & 89.9 & 87.1 & 92.3 & - & 91.6 \\
CCLM & 96.0 & 93.3 & 93.7 & 92.8 & 94.1 & 93.0 & 94.3 \\  
\midrule
\bf \babyB & 96.7 & 94.0 & 93.5 & 92.9 & 94.9 & 93.0 & 95.2 \\
\bf \babyL & \bf 97.1 & \bf 94.5 & \bf 95.1 & \bf 94.9 & \bf 95.3 & \bf 93.3 & \bf 95.6 \\
\bottomrule
\end{tabular}
}
\vspace{0.2cm}
\caption{\textbf{Results on multilingual multi-modal tasks}. All the methods except \baby rely on data that are costly to collect to perform multilingual multi-modal pre-training. We evaluate model performance in English (EN), German (DE), French (FR), Czech (CS), Chinese (ZH), and Japanese (JA). Following previous work, we report the average Recall@K for both image-to-text retrieval and text-to-image retrieval with K = 1, 5, 10. 
}
\label{tbl:mm}
\end{table}

In \baby architecture, text encoding, vision encoding and fusion are separated. Accordingly, the capabilities of vision encoding and fusion would be kept when replacing the text encoder, leading to an efficient adaptation of the new text encoder. Our study demonstrates that we can replace the text encoder after pre-training on English data with a language-specific or domain-specific one to support more applications in different languages or domains. Such a feature is hard to achieve with unified models like OFA and BEiT-3. For instance, BEiT-3 shares image, text, and fusion in a single Transformer, and thus replacing the text encoder can cause the capabilities of image encoding and fusion to be lost as well.

In this section, we replace the English text encoder of \baby with a multilingual text encoder XLM-R~\cite{conneau2020unsupervised}. Then, without a second step multilingual multi-modal pre-training, we simply finetune \baby on multilingual multi-modal downstream tasks. We choose Multi30K~\cite{young2014image} and multilingual MSCOCO~\cite{chen2015microsoft, yoshikawa2017stair, li2019coco} for evaluation since other multilingual multi-modal benchmarks such as IGLUE~\cite{bugliarello2022iglue} do not have a training set. Following previous work, we compute the average Recall@K for both image-to-text retrieval and text-to-image retrieval with K = 1, 5, 10, as the evaluation metric.

We compare \baby with SoTA multilingual multi-modal pre-training methods. $\text{M}^3\text{P}$~\cite{ni2021m3p} utilizes 101G texts covering 100 languages. UC$^2$~\cite{zhou2021uc2} translates image-text pairs in English into five different languages. MURAL~\cite{jain2021mural} collects large-scale image-text pairs in 110 languages. CCLM~\cite{cclm} utilizes parallel multilingual text pairs. All these methods rely on data that are costly to collect, while \baby relieves the multilingual multi-modal pre-training process. As shown in Table~\ref{tbl:mm}, \baby surprisingly outperforms all these methods in all six languages. The results indicate the potential of \baby being applicable to other domains or languages using a different text encoder without further pre-training.

\subsection{Ablation Study} 

\begin{table}[t]

	\centering	
\resizebox{0.8\columnwidth}{!}{%
	\begin{tabular}	{l | c  c  c  c c c }
		\toprule	 	
	  & \multicolumn{2}{c}{\bf Flickr30K} & \bf VQA & \multicolumn{2}{c}{\bf RefCOCO+} & \bf OVAD \\
	 & TR & IR & test-dev & \multicolumn{1}{c}{testA$^d$} & \multicolumn{1}{c}{testB$^d$} & mAP \\
	\midrule
    Ours &  \bf 98.0 & 89.0 & \bf 78.4 & \bf 88.6 & \bf 76.7 & \bf 27.9 \\  

        w/o \baby & 96.0 & 85.9  & 77.6 & 78.6 & 59.0 & 20.6 \\ 

	\midrule
     
     w/o multi-grained align &  96.6  & 86.2 & 77.7 & 87.3 & 75.3 & 23.1 \\
     
     w/o bbox loss &  97.4 & \bf 89.6 & 78.2  & 83.6 & 66.0 & 26.8 \\

    \midrule
    
    w/o object data &  97.2 & 86.8 & 78.1 & 88.1 & 76.5 & 26.4 \\ 
    
    w/o region data &  97.8 & 89.0 & 78.0  & 84.8 & 69.3 & 22.3 \\

		\bottomrule
	\end{tabular}
	}
 \vspace{0.2cm}

	\caption
	{
		\textbf{Ablation study} of different components in the proposed framework and different types of data utilized. 
	}
	\label{tbl:ablation}

\end{table}

We conduct an in-depth ablation study and the results are given in Table~\ref{tbl:ablation}. We describe the experimental settings in Appendix. First, we investigate the role of different components in the proposed framework and conduct an ablation of multi-grained aligning and box prediction loss respectively. It should be noted that both object and region data are utilized in these two variants. The experimental results demonstrate that multi-grained aligning is more important for the model performance than the box prediction loss in all tasks, except the visual grounding task. The box prediction loss is critical to performance on visual grounding tasks, and combining the box prediction with multi-grained aligning further improves the model performances (Ours vs. w/o bbox loss).

Second, we explore the impact of different types of annotation data used in \baby, and ablate object data and region data respectively. Both multi-grained aligning and box prediction loss are applied in these two variants. The results indicate that both types of annotations are important to performance. Object data improve image-text retrieval, while region data are critical to visual grounding and open-vocabulary attribute detection. Combining object and region data yields the best performances (Ours vs. w/o object and w/o region). The w/o \baby variant, which ablates both multi-grained aligning and box prediction loss, or both object and region data, has the worst performances in all the tasks. We also provide an ablation study on temporal modeling methods in Appendix.

\subsection{Qualitative Study of Multi-Grained Alignments}

\begin{figure*}[t]
\begin{center}
\centerline{\includegraphics[width=1\columnwidth]{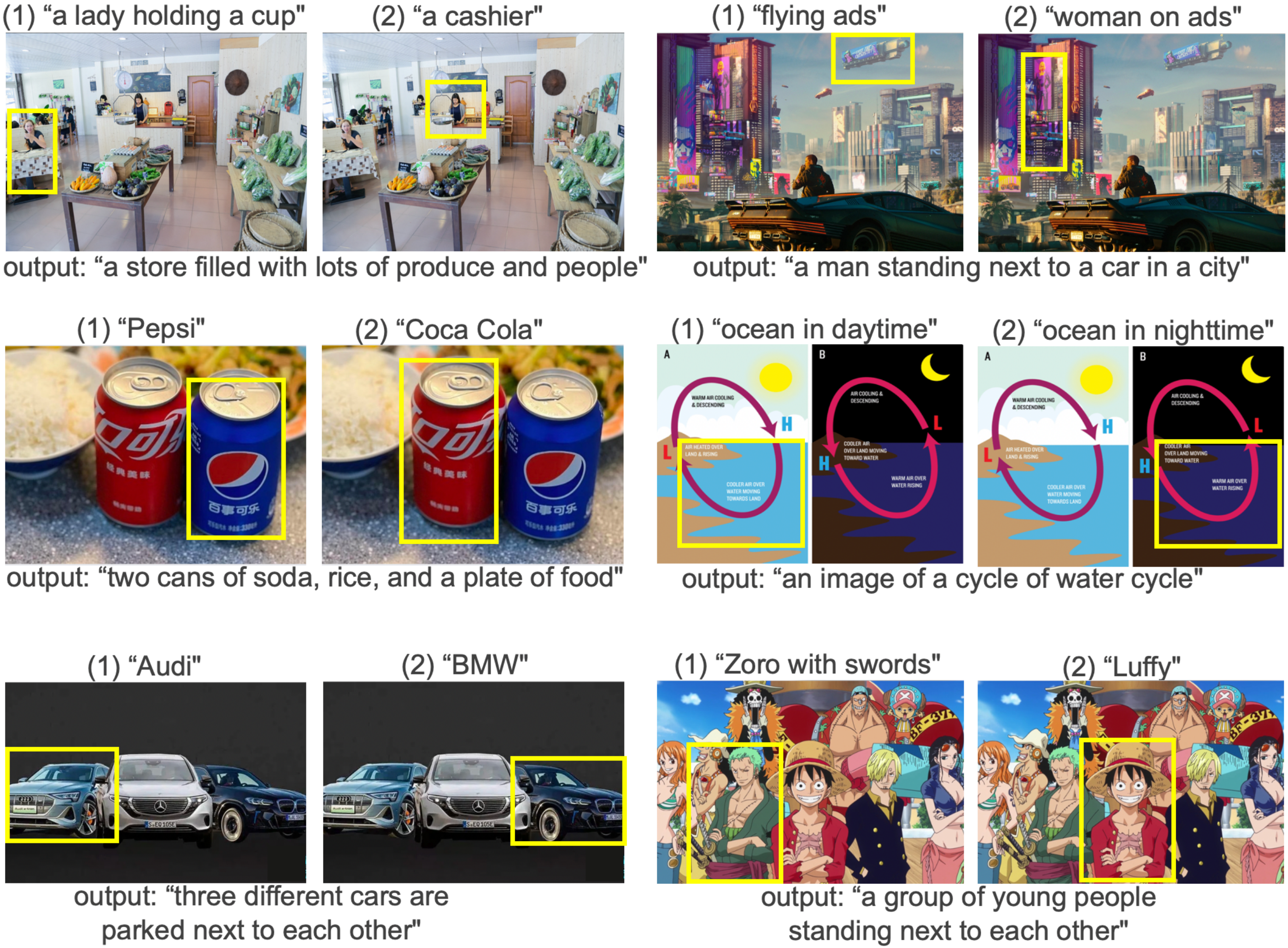}}
\caption{\textbf{Visualization of \baby generating image captions and locating visual concepts given manual input texts.} Only the image in the upper left corner is from the COCO dataset. Others are out-of-domain images from the internet. We give more examples in Appendix where we test \baby on images from robot grasping, e-commerce websites, and children's textbooks. }
\label{Fig:grounding}
\end{center}
\end{figure*}

In this section, we provide a qualitative study of what vision language alignments have been learned by \baby. To this end, we ask \baby to generate image captions to see if it can describe an image appropriately. We also ask \baby to locate visual concepts in an image given manual input descriptions to see whether it can understand fine-grained objects or regions in an image. We use \babyL fine-tuned on COCO Caption and RefCOCO+ dataset respectively for this evaluation. We visualize the results in Figure~\ref{Fig:grounding}, in which we choose some out-of-domain images from scientific posters, video games, cartoons, etc.

The visualization examples show that \baby can describe all these images appropriately with a precise understanding of the main characters or objects and their relationships. When asking \baby to locate visual concepts in an image according to the descriptions we provided, we find that it can capture small objects in the background or objects which have been partially obscured. Moreover, \baby can recognize different brands of soda or cars or distinguish ``Luffy'' and ``Zoro'' from other cartoon characters. We give more examples in Appendix, where \baby can also recognize ``Albert Einstein'', ``Edison'', ``Ultraman'', and ``Doraemon''. It is surprising since the annotations of objects or regions we exploited in pre-training are only about common objects such as ``soda'', ``car'', or ``man''. The results indicate that \baby learns to localize diverse fine-grained visual concepts from large-scale noisy image-text pairs. 

\section{Conclusion and Discussion}
In this paper, we have proposed to learn multi-grained alignments between vision and language in pre-training. To this end, we have proposed a unified framework for multi-grained vision language pre-training that directly aligns the multi-grained vision features with the paired text features and simultaneously locates multi-grained visual concepts in the same image given different text descriptions. Based on it, we have presented \baby, an all-in-one pre-trained VLM with a flexible modular architecture, in which we have further unified image encoding and video encoding to make it able to handle both image-text tasks and video-text tasks.

We have conducted extensive experiments to verify the effectiveness of \baby. The results have shown that \baby substantially outperforms SoTA image-text pre-training methods on base and large scale in many downstream image-text tasks, making a good trade-off between performance and model scale. \baby is also the new SoTA pre-trained model on video-text tasks, including video-text retrieval and video VQA. Experimental results also show that the proposed framework for multi-grained vision language pre-training is scalable to massive data and a larger model size. Moreover, we have revealed the potential of the modular design of \baby, showing it can be utilized in other languages or domains. By replacing the text encoder with XLM-R after pre-training on English data, \baby outperforms SoTA methods on multi-lingual multi-modal tasks.

We also have provided an in-depth ablation study to investigate the role of different components in the proposed framework. Experimental results have shown that both multi-grained localization and multi-grained aligning are critical components of the proposed method. Furthermore, we have conducted a qualitative study of what vision language alignments have been learned by \baby. We have found that by training with large-scale image-text pairs, \baby learns to locate diverse fine-grained visual concepts in open-domain images, such as different brands of sodas, cars, and characters or celebrities.

\section*{Acknowledgements}
We sincerely thank our colleagues at ByteDance, Tao Kong, for his constructive and detailed feedback on this work, and Jiaze Chen for his generous assistance in the training of \baby.

\bibliographystyle{unsrtnat}
\bibliography{neurips_2022}

\begin{thebibliography}{79}
\providecommand{\natexlab}[1]{#1}
\providecommand{\url}[1]{\texttt{#1}}
\expandafter\ifx\csname urlstyle\endcsname\relax
  \providecommand{\doi}[1]{doi: #1}\else
  \providecommand{\doi}{doi: \begingroup \urlstyle{rm}\Url}\fi

\bibitem[He et~al.(2016)He, Zhang, Ren, and Sun]{he2016deep}
Kaiming He, Xiangyu Zhang, Shaoqing Ren, and Jian Sun.
\newblock Deep residual learning for image recognition.
\newblock In \emph{Proceedings of the IEEE conference on computer vision and
  pattern recognition}, pages 770--778, 2016.

\bibitem[Dosovitskiy et~al.(2020)Dosovitskiy, Beyer, Kolesnikov, Weissenborn,
  Zhai, Unterthiner, Dehghani, Minderer, Heigold, Gelly,
  et~al.]{dosovitskiy2020image}
Alexey Dosovitskiy, Lucas Beyer, Alexander Kolesnikov, Dirk Weissenborn,
  Xiaohua Zhai, Thomas Unterthiner, Mostafa Dehghani, Matthias Minderer, Georg
  Heigold, Sylvain Gelly, et~al.
\newblock An image is worth 16x16 words: Transformers for image recognition at
  scale.
\newblock In \emph{International Conference on Learning Representations}, 2020.

\bibitem[Huang et~al.(2020)Huang, Zeng, Liu, Fu, and Fu]{huang2020pixel}
Zhicheng Huang, Zhaoyang Zeng, Bei Liu, Dongmei Fu, and Jianlong Fu.
\newblock Pixel-bert: Aligning image pixels with text by deep multi-modal
  transformers.
\newblock \emph{arXiv preprint arXiv:2004.00849}, 2020.

\bibitem[Kim et~al.(2021)Kim, Son, and Kim]{kim2021vilt}
Wonjae Kim, Bokyung Son, and Ildoo Kim.
\newblock Vilt: Vision-and-language transformer without convolution or region
  supervision.
\newblock In \emph{International Conference on Machine Learning}, pages
  5583--5594. PMLR, 2021.

\bibitem[Li et~al.(2021)Li, Selvaraju, Gotmare, Joty, Xiong, and
  Hoi]{li2021align}
Junnan Li, Ramprasaath Selvaraju, Akhilesh Gotmare, Shafiq Joty, Caiming Xiong,
  and Steven Chu~Hong Hoi.
\newblock Align before fuse: Vision and language representation learning with
  momentum distillation.
\newblock \emph{Advances in Neural Information Processing Systems}, 34, 2021.

\bibitem[Huo et~al.(2021)Huo, Zhang, Liu, Lu, Gao, Yang, Wen, Zhang, Xu, Zheng,
  et~al.]{huo2021wenlan}
Yuqi Huo, Manli Zhang, Guangzhen Liu, Haoyu Lu, Yizhao Gao, Guoxing Yang,
  Jingyuan Wen, Heng Zhang, Baogui Xu, Weihao Zheng, et~al.
\newblock Wenlan: Bridging vision and language by large-scale multi-modal
  pre-training.
\newblock \emph{arXiv preprint arXiv:2103.06561}, 2021.

\bibitem[Tan and Bansal(2019)]{tan2019lxmert}
Hao Tan and Mohit Bansal.
\newblock {LXMERT}: Learning cross-modality encoder representations from
  transformers.
\newblock In \emph{Proceedings of the 2019 Conference on Empirical Methods in
  Natural Language Processing and the 9th International Joint Conference on
  Natural Language Processing (EMNLP-IJCNLP)}, pages 5100--5111, Hong Kong,
  China, 2019. Association for Computational Linguistics.
\newblock \doi{10.18653/v1/D19-1514}.

\bibitem[Lu et~al.(2019)Lu, Batra, Parikh, and Lee]{lu2019vilbert}
Jiasen Lu, Dhruv Batra, Devi Parikh, and Stefan Lee.
\newblock Vilbert: Pretraining task-agnostic visiolinguistic representations
  for vision-and-language tasks.
\newblock In \emph{Advances in Neural Information Processing Systems 32: Annual
  Conference on Neural Information Processing Systems 2019, NeurIPS 2019,
  December 8-14, 2019, Vancouver, BC, Canada}, pages 13--23, 2019.

\bibitem[Li et~al.(2019{\natexlab{a}})Li, Yatskar, Yin, Hsieh, and
  Chang]{li2019visualbert}
Liunian~Harold Li, Mark Yatskar, Da~Yin, Cho-Jui Hsieh, and Kai-Wei Chang.
\newblock Visualbert: A simple and performant baseline for vision and language.
\newblock \emph{arXiv preprint arXiv:1908.03557}, 2019{\natexlab{a}}.

\bibitem[Gan et~al.(2020)Gan, Chen, Li, Zhu, Cheng, and Liu]{gan2020large}
Zhe Gan, Yen{-}Chun Chen, Linjie Li, Chen Zhu, Yu~Cheng, and Jingjing Liu.
\newblock Large-scale adversarial training for vision-and-language
  representation learning.
\newblock In Hugo Larochelle, Marc'Aurelio Ranzato, Raia Hadsell,
  Maria{-}Florina Balcan, and Hsuan{-}Tien Lin, editors, \emph{Advances in
  Neural Information Processing Systems 33: Annual Conference on Neural
  Information Processing Systems 2020, NeurIPS 2020, December 6-12, 2020,
  virtual}, 2020.

\bibitem[Chen et~al.(2020)Chen, Li, Yu, El~Kholy, Ahmed, Gan, Cheng, and
  Liu]{chen2020uniter}
Yen-Chun Chen, Linjie Li, Licheng Yu, Ahmed El~Kholy, Faisal Ahmed, Zhe Gan,
  Yu~Cheng, and Jingjing Liu.
\newblock Uniter: Universal image-text representation learning.
\newblock In \emph{European conference on computer vision}, pages 104--120.
  Springer, 2020.

\bibitem[Li et~al.(2020)Li, Yin, Li, Zhang, Hu, Zhang, Wang, Hu, Dong, Wei,
  et~al.]{li2020oscar}
Xiujun Li, Xi~Yin, Chunyuan Li, Pengchuan Zhang, Xiaowei Hu, Lei Zhang, Lijuan
  Wang, Houdong Hu, Li~Dong, Furu Wei, et~al.
\newblock Oscar: Object-semantics aligned pre-training for vision-language
  tasks.
\newblock In \emph{European Conference on Computer Vision}, pages 121--137.
  Springer, 2020.

\bibitem[Zhang et~al.(2021)Zhang, Li, Hu, Yang, Zhang, Wang, Choi, and
  Gao]{zhang2021vinvl}
Pengchuan Zhang, Xiujun Li, Xiaowei Hu, Jianwei Yang, Lei Zhang, Lijuan Wang,
  Yejin Choi, and Jianfeng Gao.
\newblock Vinvl: Revisiting visual representations in vision-language models.
\newblock In \emph{Proceedings of the IEEE/CVF Conference on Computer Vision
  and Pattern Recognition}, pages 5579--5588, 2021.

\bibitem[Lin et~al.(2014)Lin, Maire, Belongie, Hays, Perona, Ramanan,
  Doll{\'a}r, and Zitnick]{lin2014microsoft}
Tsung-Yi Lin, Michael Maire, Serge Belongie, James Hays, Pietro Perona, Deva
  Ramanan, Piotr Doll{\'a}r, and C~Lawrence Zitnick.
\newblock Microsoft coco: Common objects in context.
\newblock In \emph{European conference on computer vision}, pages 740--755.
  Springer, 2014.

\bibitem[Shao et~al.(2019)Shao, Li, Zhang, Peng, Yu, Zhang, Li, and
  Sun]{shao2019objects365}
Shuai Shao, Zeming Li, Tianyuan Zhang, Chao Peng, Gang Yu, Xiangyu Zhang, Jing
  Li, and Jian Sun.
\newblock Objects365: {A} large-scale, high-quality dataset for object
  detection.
\newblock In \emph{2019 {IEEE/CVF} International Conference on Computer Vision,
  {ICCV} 2019, Seoul, Korea (South), October 27 - November 2, 2019}, pages
  8429--8438. {IEEE}, 2019.
\newblock \doi{10.1109/ICCV.2019.00852}.

\bibitem[Kuznetsova et~al.(2018)Kuznetsova, Rom, Alldrin, Uijlings, Krasin,
  Pont-Tuset, Kamali, Popov, Malloci, Kolesnikov, et~al.]{kuznetsova2018open}
Alina Kuznetsova, Hassan Rom, Neil Alldrin, Jasper Uijlings, Ivan Krasin, Jordi
  Pont-Tuset, Shahab Kamali, Stefan Popov, Matteo Malloci, Alexander
  Kolesnikov, et~al.
\newblock The open images dataset v4: Unified image classification, object
  detection, and visual relationship detection at scale.
\newblock \emph{arXiv preprint arXiv:1811.00982}, 2018.

\bibitem[Krishna et~al.(2017)Krishna, Zhu, Groth, Johnson, Hata, Kravitz, Chen,
  Kalantidis, Li, Shamma, et~al.]{krishna2016visual}
Ranjay Krishna, Yuke Zhu, Oliver Groth, Justin Johnson, Kenji Hata, Joshua
  Kravitz, Stephanie Chen, Yannis Kalantidis, Li-Jia Li, David~A Shamma, et~al.
\newblock Visual genome: Connecting language and vision using crowdsourced
  dense image annotations.
\newblock \emph{International journal of computer vision}, 123\penalty0
  (1):\penalty0 32--73, 2017.

\bibitem[Vaswani et~al.(2017)Vaswani, Shazeer, Parmar, Uszkoreit, Jones, Gomez,
  Kaiser, and Polosukhin]{vaswani2017attention}
Ashish Vaswani, Noam Shazeer, Niki Parmar, Jakob Uszkoreit, Llion Jones,
  Aidan~N. Gomez, Lukasz Kaiser, and Illia Polosukhin.
\newblock Attention is all you need.
\newblock In Isabelle Guyon, Ulrike von Luxburg, Samy Bengio, Hanna~M. Wallach,
  Rob Fergus, S.~V.~N. Vishwanathan, and Roman Garnett, editors, \emph{Advances
  in Neural Information Processing Systems 30: Annual Conference on Neural
  Information Processing Systems 2017, December 4-9, 2017, Long Beach, CA,
  {USA}}, pages 5998--6008, 2017.

\bibitem[Wang et~al.(2021{\natexlab{a}})Wang, Yu, Yu, Dai, Tsvetkov, and
  Cao]{wang2021simvlm}
Zirui Wang, Jiahui Yu, Adams~Wei Yu, Zihang Dai, Yulia Tsvetkov, and Yuan Cao.
\newblock Simvlm: Simple visual language model pretraining with weak
  supervision.
\newblock \emph{arXiv preprint arXiv:2108.10904}, 2021{\natexlab{a}}.

\bibitem[Li et~al.(2022{\natexlab{a}})Li, Li, Xiong, and Hoi]{li2022blip}
Junnan Li, Dongxu Li, Caiming Xiong, and Steven Hoi.
\newblock Blip: Bootstrapping language-image pre-training for unified
  vision-language understanding and generation.
\newblock \emph{arXiv preprint arXiv:2201.12086}, 2022{\natexlab{a}}.

\bibitem[Kamath et~al.(2021)Kamath, Singh, LeCun, Synnaeve, Misra, and
  Carion]{kamath2021mdetr}
Aishwarya Kamath, Mannat Singh, Yann LeCun, Gabriel Synnaeve, Ishan Misra, and
  Nicolas Carion.
\newblock Mdetr-modulated detection for end-to-end multi-modal understanding.
\newblock In \emph{Proceedings of the IEEE/CVF International Conference on
  Computer Vision}, pages 1780--1790, 2021.

\bibitem[Wang et~al.(2022{\natexlab{a}})Wang, Yang, Men, Lin, Bai, Li, Ma,
  Zhou, Zhou, and Yang]{wang2022ofa}
Peng Wang, An~Yang, Rui Men, Junyang Lin, Shuai Bai, Zhikang Li, Jianxin Ma,
  Chang Zhou, Jingren Zhou, and Hongxia Yang.
\newblock Ofa: Unifying architectures, tasks, and modalities through a simple
  sequence-to-sequence learning framework.
\newblock In \emph{International Conference on Machine Learning}, pages
  23318--23340. PMLR, 2022{\natexlab{a}}.

\bibitem[Yu et~al.(2022)Yu, Wang, Vasudevan, Yeung, Seyedhosseini, and
  Wu]{yu2022coca}
Jiahui Yu, Zirui Wang, Vijay Vasudevan, Legg Yeung, Mojtaba Seyedhosseini, and
  Yonghui Wu.
\newblock Coca: Contrastive captioners are image-text foundation models.
\newblock \emph{arXiv preprint arXiv:2205.01917}, 2022.

\bibitem[Wang et~al.(2022{\natexlab{b}})Wang, Bao, Dong, Bjorck, Peng, Liu,
  Aggarwal, Mohammed, Singhal, Som, et~al.]{wang2022image}
Wenhui Wang, Hangbo Bao, Li~Dong, Johan Bjorck, Zhiliang Peng, Qiang Liu, Kriti
  Aggarwal, Owais~Khan Mohammed, Saksham Singhal, Subhojit Som, et~al.
\newblock Image as a foreign language: Beit pretraining for all vision and
  vision-language tasks.
\newblock \emph{arXiv preprint arXiv:2208.10442}, 2022{\natexlab{b}}.

\bibitem[Conneau et~al.(2020)Conneau, Khandelwal, Goyal, Chaudhary, Wenzek,
  Guzm{\'a}n, Grave, Ott, Zettlemoyer, and Stoyanov]{conneau2020unsupervised}
Alexis Conneau, Kartikay Khandelwal, Naman Goyal, Vishrav Chaudhary, Guillaume
  Wenzek, Francisco Guzm{\'a}n, {\'E}douard Grave, Myle Ott, Luke Zettlemoyer,
  and Veselin Stoyanov.
\newblock Unsupervised cross-lingual representation learning at scale.
\newblock In \emph{Proceedings of the 58th Annual Meeting of the Association
  for Computational Linguistics}, pages 8440--8451, 2020.

\bibitem[Zhou et~al.(2021)Zhou, Zhou, Wang, Cheng, Li, Yu, and
  Liu]{zhou2021uc2}
Mingyang Zhou, Luowei Zhou, Shuohang Wang, Yu~Cheng, Linjie Li, Zhou Yu, and
  Jingjing Liu.
\newblock Uc2: Universal cross-lingual cross-modal vision-and-language
  pre-training.
\newblock In \emph{Proceedings of the IEEE/CVF Conference on Computer Vision
  and Pattern Recognition}, pages 4155--4165, 2021.

\bibitem[Jain et~al.(2021)Jain, Guo, Srinivasan, Chen, Kudugunta, Jia, Yang,
  and Baldridge]{jain2021mural}
Aashi Jain, Mandy Guo, Krishna Srinivasan, Ting Chen, Sneha Kudugunta, Chao
  Jia, Yinfei Yang, and Jason Baldridge.
\newblock Mural: multimodal, multitask retrieval across languages.
\newblock \emph{ArXiv preprint}, abs/2109.05125, 2021.

\bibitem[Zeng et~al.(2022)Zeng, Zhou, Luo, and Zhang]{cclm}
Yan Zeng, Wangchunshu Zhou, Ao~Luo, and Xinsong Zhang.
\newblock Cross-view language modeling: Towards unified cross-lingual
  cross-modal pre-training.
\newblock \emph{arXiv preprint arXiv:2206.00621}, 2022.

\bibitem[Ren et~al.(2015)Ren, He, Girshick, and Sun]{ren2015faster}
Shaoqing Ren, Kaiming He, Ross~B. Girshick, and Jian Sun.
\newblock Faster {R-CNN:} towards real-time object detection with region
  proposal networks.
\newblock In Corinna Cortes, Neil~D. Lawrence, Daniel~D. Lee, Masashi Sugiyama,
  and Roman Garnett, editors, \emph{Advances in Neural Information Processing
  Systems 28: Annual Conference on Neural Information Processing Systems 2015,
  December 7-12, 2015, Montreal, Quebec, Canada}, pages 91--99, 2015.

\bibitem[Anderson et~al.(2018)Anderson, He, Buehler, Teney, Johnson, Gould, and
  Zhang]{anderson2018bottom}
Peter Anderson, Xiaodong He, Chris Buehler, Damien Teney, Mark Johnson, Stephen
  Gould, and Lei Zhang.
\newblock Bottom-up and top-down attention for image captioning and visual
  question answering.
\newblock In \emph{2018 {IEEE} Conference on Computer Vision and Pattern
  Recognition, {CVPR} 2018, Salt Lake City, UT, USA, June 18-22, 2018}, pages
  6077--6086. {IEEE} Computer Society, 2018.
\newblock \doi{10.1109/CVPR.2018.00636}.

\bibitem[Jiang et~al.(2020)Jiang, Misra, Rohrbach, Learned-Miller, and
  Chen]{jiang2020defense}
Huaizu Jiang, Ishan Misra, Marcus Rohrbach, Erik Learned-Miller, and Xinlei
  Chen.
\newblock In defense of grid features for visual question answering.
\newblock In \emph{Proceedings of the IEEE/CVF Conference on Computer Vision
  and Pattern Recognition}, pages 10267--10276, 2020.

\bibitem[Liu et~al.(2021{\natexlab{a}})Liu, Lin, Cao, Hu, Wei, Zhang, Lin, and
  Guo]{liu2021swin}
Ze~Liu, Yutong Lin, Yue Cao, Han Hu, Yixuan Wei, Zheng Zhang, Stephen Lin, and
  Baining Guo.
\newblock Swin transformer: Hierarchical vision transformer using shifted
  windows.
\newblock \emph{arXiv preprint arXiv:2103.14030}, 2021{\natexlab{a}}.

\bibitem[Peng et~al.(2022)Peng, Dong, Bao, Ye, and Wei]{peng2022beit}
Zhiliang Peng, Li~Dong, Hangbo Bao, Qixiang Ye, and Furu Wei.
\newblock Beit v2: Masked image modeling with vector-quantized visual
  tokenizers.
\newblock \emph{arXiv preprint arXiv:2208.06366}, 2022.

\bibitem[Dou et~al.(2021)Dou, Xu, Gan, Wang, Wang, Wang, Zhu, Liu, Zeng,
  et~al.]{dou2021empirical}
Zi-Yi Dou, Yichong Xu, Zhe Gan, Jianfeng Wang, Shuohang Wang, Lijuan Wang,
  Chenguang Zhu, Zicheng Liu, Michael Zeng, et~al.
\newblock An empirical study of training end-to-end vision-and-language
  transformers.
\newblock \emph{arXiv preprint arXiv:2111.02387}, 2021.

\bibitem[Bao et~al.(2022)Bao, Wang, Dong, and Wei]{bao2022vl}
Hangbo Bao, Wenhui Wang, Li~Dong, and Furu Wei.
\newblock Vl-beit: Generative vision-language pretraining.
\newblock \emph{arXiv preprint arXiv:2206.01127}, 2022.

\bibitem[Xu et~al.(2021)Xu, Yan, Li, Bi, Huang, Xiao, and Huang]{xu2021e2e}
Haiyang Xu, Ming Yan, Chenliang Li, Bin Bi, Songfang Huang, Wenming Xiao, and
  Fei Huang.
\newblock {E}2{E}-{VLP}: End-to-end vision-language pre-training enhanced by
  visual learning.
\newblock In \emph{Proceedings of the 59th Annual Meeting of the Association
  for Computational Linguistics and the 11th International Joint Conference on
  Natural Language Processing (Volume 1: Long Papers)}, pages 503--513, Online,
  2021. Association for Computational Linguistics.
\newblock \doi{10.18653/v1/2021.acl-long.42}.

\bibitem[Carion et~al.(2020)Carion, Massa, Synnaeve, Usunier, Kirillov, and
  Zagoruyko]{carion2020end}
Nicolas Carion, Francisco Massa, Gabriel Synnaeve, Nicolas Usunier, Alexander
  Kirillov, and Sergey Zagoruyko.
\newblock End-to-end object detection with transformers.
\newblock In \emph{European Conference on Computer Vision}, pages 213--229.
  Springer, 2020.

\bibitem[Liu et~al.(2021{\natexlab{b}})Liu, Wu, Tseng, Lal, He, and
  Duan]{liu2021kd}
Yongfei Liu, Chenfei Wu, Shao-yen Tseng, Vasudev Lal, Xuming He, and Nan Duan.
\newblock Kd-vlp: Improving end-to-end vision-and-language pretraining with
  object knowledge distillation.
\newblock \emph{arXiv preprint arXiv:2109.10504}, 2021{\natexlab{b}}.

\bibitem[Lei et~al.(2021)Lei, Li, Zhou, Gan, Berg, Bansal, and
  Liu]{lei2021less}
Jie Lei, Linjie Li, Luowei Zhou, Zhe Gan, Tamara~L Berg, Mohit Bansal, and
  Jingjing Liu.
\newblock Less is more: Clipbert for video-and-language learning via sparse
  sampling.
\newblock In \emph{Proceedings of the IEEE/CVF Conference on Computer Vision
  and Pattern Recognition}, pages 7331--7341, 2021.

\bibitem[Bain et~al.(2021)Bain, Nagrani, Varol, and Zisserman]{bain2021frozen}
Max Bain, Arsha Nagrani, G{\"u}l Varol, and Andrew Zisserman.
\newblock Frozen in time: A joint video and image encoder for end-to-end
  retrieval.
\newblock In \emph{Proceedings of the IEEE/CVF International Conference on
  Computer Vision}, pages 1728--1738, 2021.

\bibitem[Li et~al.(2022{\natexlab{b}})Li, Li, Li, Niebles, and
  Hoi]{li2022align}
Dongxu Li, Junnan Li, Hongdong Li, Juan~Carlos Niebles, and Steven~CH Hoi.
\newblock Align and prompt: Video-and-language pre-training with entity
  prompts.
\newblock In \emph{Proceedings of the IEEE/CVF Conference on Computer Vision
  and Pattern Recognition}, pages 4953--4963, 2022{\natexlab{b}}.

\bibitem[Fu et~al.(2021)Fu, Li, Gan, Lin, Wang, Wang, and Liu]{fu2021violet}
Tsu-Jui Fu, Linjie Li, Zhe Gan, Kevin Lin, William~Yang Wang, Lijuan Wang, and
  Zicheng Liu.
\newblock Violet: End-to-end video-language transformers with masked
  visual-token modeling.
\newblock \emph{arXiv preprint arXiv:2111.12681}, 2021.

\bibitem[Wang et~al.(2022{\natexlab{c}})Wang, Ge, Yan, Ge, Lin, Cai, Wu, Shan,
  Qie, and Shou]{wang2022all}
Alex~Jinpeng Wang, Yixiao Ge, Rui Yan, Yuying Ge, Xudong Lin, Guanyu Cai,
  Jianping Wu, Ying Shan, Xiaohu Qie, and Mike~Zheng Shou.
\newblock All in one: Exploring unified video-language pre-training.
\newblock \emph{arXiv preprint arXiv:2203.07303}, 2022{\natexlab{c}}.

\bibitem[Xue et~al.(2022)Xue, Sun, Liu, Fu, Song, Li, and Luo]{xue2022clip}
Hongwei Xue, Yuchong Sun, Bei Liu, Jianlong Fu, Ruihua Song, Houqiang Li, and
  Jiebo Luo.
\newblock Clip-vip: Adapting pre-trained image-text model to video-language
  representation alignment.
\newblock \emph{arXiv preprint arXiv:2209.06430}, 2022.

\bibitem[Min et~al.(2022)Min, Kong, Tu, Gong, Cai, Zhao, Liu, Zheng, Wang, Li,
  et~al.]{min2022hunyuan_tvr}
Shaobo Min, Weijie Kong, Rong-Cheng Tu, Dihong Gong, Chengfei Cai, Wenzhe Zhao,
  Chenyang Liu, Sixiao Zheng, Hongfa Wang, Zhifeng Li, et~al.
\newblock Hunyuan\_tvr for text-video retrivial.
\newblock \emph{arXiv preprint arXiv:2204.03382}, 2022.

\bibitem[Yang et~al.(2022)Yang, Miech, Sivic, Laptev, and Schmid]{yang2022zero}
Antoine Yang, Antoine Miech, Josef Sivic, Ivan Laptev, and Cordelia Schmid.
\newblock Zero-shot video question answering via frozen bidirectional language
  models.
\newblock \emph{arXiv preprint arXiv:2206.08155}, 2022.

\bibitem[Wang et~al.(2022{\natexlab{d}})Wang, Chen, Wu, Luo, Zhou, Zhao, Xie,
  Liu, Jiang, and Yuan]{wang2022omnivl}
Junke Wang, Dongdong Chen, Zuxuan Wu, Chong Luo, Luowei Zhou, Yucheng Zhao,
  Yujia Xie, Ce~Liu, Yu-Gang Jiang, and Lu~Yuan.
\newblock Omnivl: One foundation model for image-language and video-language
  tasks.
\newblock \emph{arXiv preprint arXiv:2209.07526}, 2022{\natexlab{d}}.

\bibitem[Bertasius et~al.(2021)Bertasius, Wang, and
  Torresani]{bertasius2021space}
Gedas Bertasius, Heng Wang, and Lorenzo Torresani.
\newblock Is space-time attention all you need for video understanding?
\newblock In \emph{ICML}, volume~2, page~4, 2021.

\bibitem[Ni et~al.(2021)Ni, Huang, Su, Cui, Bharti, Wang, Zhang, and
  Duan]{ni2021m3p}
Minheng Ni, Haoyang Huang, Lin Su, Edward Cui, Taroon Bharti, Lijuan Wang,
  Dongdong Zhang, and Nan Duan.
\newblock M3p: Learning universal representations via multitask multilingual
  multimodal pre-training.
\newblock In \emph{Proceedings of the IEEE/CVF Conference on Computer Vision
  and Pattern Recognition}, pages 3977--3986, 2021.

\bibitem[Radford et~al.(2021)Radford, Kim, Hallacy, Ramesh, Goh, Agarwal,
  Sastry, Askell, Mishkin, Clark, Krueger, and Sutskever]{radford2021learning}
Alec Radford, Jong~Wook Kim, Chris Hallacy, Aditya Ramesh, Gabriel Goh,
  Sandhini Agarwal, Girish Sastry, Amanda Askell, Pamela Mishkin, Jack Clark,
  Gretchen Krueger, and Ilya Sutskever.
\newblock Learning transferable visual models from natural language
  supervision.
\newblock In Marina Meila and Tong Zhang, editors, \emph{Proceedings of the
  38th International Conference on Machine Learning, {ICML} 2021, 18-24 July
  2021, Virtual Event}, volume 139 of \emph{Proceedings of Machine Learning
  Research}, pages 8748--8763. {PMLR}, 2021.

\bibitem[Rezatofighi et~al.(2019)Rezatofighi, Tsoi, Gwak, Sadeghian, Reid, and
  Savarese]{rezatofighi2019generalized}
Hamid Rezatofighi, Nathan Tsoi, JunYoung Gwak, Amir Sadeghian, Ian~D. Reid, and
  Silvio Savarese.
\newblock Generalized intersection over union: {A} metric and a loss for
  bounding box regression.
\newblock In \emph{{IEEE} Conference on Computer Vision and Pattern
  Recognition, {CVPR} 2019, Long Beach, CA, USA, June 16-20, 2019}, pages
  658--666. Computer Vision Foundation / {IEEE}, 2019.
\newblock \doi{10.1109/CVPR.2019.00075}.

\bibitem[Ordonez et~al.(2011)Ordonez, Kulkarni, and Berg]{ordonez2011im2text}
Vicente Ordonez, Girish Kulkarni, and Tamara~L. Berg.
\newblock Im2text: Describing images using 1 million captioned photographs.
\newblock In John Shawe{-}Taylor, Richard~S. Zemel, Peter~L. Bartlett, Fernando
  C.~N. Pereira, and Kilian~Q. Weinberger, editors, \emph{Advances in Neural
  Information Processing Systems 24: 25th Annual Conference on Neural
  Information Processing Systems 2011. Proceedings of a meeting held 12-14
  December 2011, Granada, Spain}, pages 1143--1151, 2011.

\bibitem[Sharma et~al.(2018)Sharma, Ding, Goodman, and
  Soricut]{sharma2018conceptual}
Piyush Sharma, Nan Ding, Sebastian Goodman, and Radu Soricut.
\newblock Conceptual captions: A cleaned, hypernymed, image alt-text dataset
  for automatic image captioning.
\newblock In \emph{Proceedings of the 56th Annual Meeting of the Association
  for Computational Linguistics (Volume 1: Long Papers)}, pages 2556--2565,
  Melbourne, Australia, 2018. Association for Computational Linguistics.
\newblock \doi{10.18653/v1/P18-1238}.

\bibitem[Yu et~al.(2016)Yu, Poirson, Yang, Berg, and Berg]{yu2016modeling}
Licheng Yu, Patrick Poirson, Shan Yang, Alexander~C Berg, and Tamara~L Berg.
\newblock Modeling context in referring expressions.
\newblock In \emph{European Conference on Computer Vision}, pages 69--85.
  Springer, 2016.

\bibitem[Hudson and Manning(2019)]{hudson2019gqa}
Drew~A Hudson and Christopher~D Manning.
\newblock Gqa: A new dataset for real-world visual reasoning and compositional
  question answering.
\newblock In \emph{Proceedings of the IEEE/CVF conference on computer vision
  and pattern recognition}, pages 6700--6709, 2019.

\bibitem[Plummer et~al.(2015)Plummer, Wang, Cervantes, Caicedo, Hockenmaier,
  and Lazebnik]{plummer2015flickr30k}
Bryan~A. Plummer, Liwei Wang, Chris~M. Cervantes, Juan~C. Caicedo, Julia
  Hockenmaier, and Svetlana Lazebnik.
\newblock Flickr30k entities: Collecting region-to-phrase correspondences for
  richer image-to-sentence models.
\newblock In \emph{2015 {IEEE} International Conference on Computer Vision,
  {ICCV} 2015, Santiago, Chile, December 7-13, 2015}, pages 2641--2649. {IEEE}
  Computer Society, 2015.
\newblock \doi{10.1109/ICCV.2015.303}.

\bibitem[Changpinyo et~al.(2021)Changpinyo, Sharma, Ding, and
  Soricut]{changpinyo2021conceptual}
Soravit Changpinyo, Piyush Sharma, Nan Ding, and Radu Soricut.
\newblock Conceptual 12m: Pushing web-scale image-text pre-training to
  recognize long-tail visual concepts.
\newblock In \emph{Proceedings of the IEEE/CVF Conference on Computer Vision
  and Pattern Recognition}, pages 3558--3568, 2021.

\bibitem[Schuhmann et~al.(2022)Schuhmann, Beaumont, Vencu, Gordon, Wightman,
  Cherti, Coombes, Katta, Mullis, Wortsman, et~al.]{schuhmann2022laion}
Christoph Schuhmann, Romain Beaumont, Richard Vencu, Cade Gordon, Ross
  Wightman, Mehdi Cherti, Theo Coombes, Aarush Katta, Clayton Mullis, Mitchell
  Wortsman, et~al.
\newblock Laion-5b: An open large-scale dataset for training next generation
  image-text models.
\newblock \emph{arXiv preprint arXiv:2210.08402}, 2022.

\bibitem[Miech et~al.(2019)Miech, Zhukov, Alayrac, Tapaswi, Laptev, and
  Sivic]{miech2019howto100m}
Antoine Miech, Dimitri Zhukov, Jean-Baptiste Alayrac, Makarand Tapaswi, Ivan
  Laptev, and Josef Sivic.
\newblock Howto100m: Learning a text-video embedding by watching hundred
  million narrated video clips.
\newblock In \emph{Proceedings of the IEEE/CVF International Conference on
  Computer Vision}, pages 2630--2640, 2019.

\bibitem[Zellers et~al.(2021)Zellers, Lu, Hessel, Yu, Park, Cao, Farhadi, and
  Choi]{zellers2021merlot}
Rowan Zellers, Ximing Lu, Jack Hessel, Youngjae Yu, Jae~Sung Park, Jize Cao,
  Ali Farhadi, and Yejin Choi.
\newblock Merlot: Multimodal neural script knowledge models.
\newblock \emph{Advances in Neural Information Processing Systems},
  34:\penalty0 23634--23651, 2021.

\bibitem[Wang et~al.(2022{\natexlab{e}})Wang, Zhou, Zeng, and
  Zhang]{wang2022efficientvlm}
Tiannan Wang, Wangchunshu Zhou, Yan Zeng, and Xinsong Zhang.
\newblock Efficientvlm: Fast and accurate vision-language models via knowledge
  distillation and modal-adaptive pruning.
\newblock \emph{arXiv preprint arXiv:2210.07795}, 2022{\natexlab{e}}.

\bibitem[Devlin et~al.(2019)Devlin, Chang, Lee, and Toutanova]{devlin2019bert}
Jacob Devlin, Ming-Wei Chang, Kenton Lee, and Kristina Toutanova.
\newblock {BERT}: Pre-training of deep bidirectional transformers for language
  understanding.
\newblock In \emph{Proceedings of the 2019 Conference of the North {A}merican
  Chapter of the Association for Computational Linguistics: Human Language
  Technologies, Volume 1 (Long and Short Papers)}, pages 4171--4186,
  Minneapolis, Minnesota, 2019. Association for Computational Linguistics.
\newblock \doi{10.18653/v1/N19-1423}.

\bibitem[Karpathy and Li(2015)]{karpathy2015deep}
Andrej Karpathy and Fei{-}Fei Li.
\newblock Deep visual-semantic alignments for generating image descriptions.
\newblock In \emph{{IEEE} Conference on Computer Vision and Pattern
  Recognition, {CVPR} 2015, Boston, MA, USA, June 7-12, 2015}, pages
  3128--3137. {IEEE} Computer Society, 2015.
\newblock \doi{10.1109/CVPR.2015.7298932}.

\bibitem[Yao et~al.(2021)Yao, Huang, Hou, Lu, Niu, Xu, Liang, Li, Jiang, and
  Xu]{yao2021filip}
Lewei Yao, Runhui Huang, Lu~Hou, Guansong Lu, Minzhe Niu, Hang Xu, Xiaodan
  Liang, Zhenguo Li, Xin Jiang, and Chunjing Xu.
\newblock Filip: Fine-grained interactive language-image pre-training.
\newblock \emph{arXiv preprint arXiv:2111.07783}, 2021.

\bibitem[Goyal et~al.(2017)Goyal, Khot, Summers{-}Stay, Batra, and
  Parikh]{goyal2017making}
Yash Goyal, Tejas Khot, Douglas Summers{-}Stay, Dhruv Batra, and Devi Parikh.
\newblock Making the {V} in {VQA} matter: Elevating the role of image
  understanding in visual question answering.
\newblock In \emph{2017 {IEEE} Conference on Computer Vision and Pattern
  Recognition, {CVPR} 2017, Honolulu, HI, USA, July 21-26, 2017}, pages
  6325--6334. {IEEE} Computer Society, 2017.
\newblock \doi{10.1109/CVPR.2017.670}.

\bibitem[Suhr et~al.(2019)Suhr, Zhou, Zhang, Zhang, Bai, and
  Artzi]{suhr2018corpus}
Alane Suhr, Stephanie Zhou, Ally Zhang, Iris Zhang, Huajun Bai, and Yoav Artzi.
\newblock A corpus for reasoning about natural language grounded in
  photographs.
\newblock In \emph{Proceedings of the 57th Annual Meeting of the Association
  for Computational Linguistics}, pages 6418--6428, Florence, Italy, 2019.
  Association for Computational Linguistics.
\newblock \doi{10.18653/v1/P19-1644}.

\bibitem[Wang et~al.(2021{\natexlab{b}})Wang, Bao, Dong, and Wei]{wang2021vlmo}
Wenhui Wang, Hangbo Bao, Li~Dong, and Furu Wei.
\newblock Vlmo: Unified vision-language pre-training with
  mixture-of-modality-experts.
\newblock \emph{arXiv preprint arXiv:2111.02358}, 2021{\natexlab{b}}.

\bibitem[Chen et~al.(2015)Chen, Fang, Lin, Vedantam, Gupta, Doll{\'a}r, and
  Zitnick]{chen2015microsoft}
Xinlei Chen, Hao Fang, Tsung-Yi Lin, Ramakrishna Vedantam, Saurabh Gupta, Piotr
  Doll{\'a}r, and C~Lawrence Zitnick.
\newblock Microsoft coco captions: Data collection and evaluation server.
\newblock \emph{arXiv preprint arXiv:1504.00325}, 2015.

\bibitem[Dong et~al.(2019)Dong, Yang, Wang, Wei, Liu, Wang, Gao, Zhou, and
  Hon]{dong2019unified}
Li~Dong, Nan Yang, Wenhui Wang, Furu Wei, Xiaodong Liu, Yu~Wang, Jianfeng Gao,
  Ming Zhou, and Hsiao-Wuen Hon.
\newblock Unified language model pre-training for natural language
  understanding and generation.
\newblock \emph{Advances in Neural Information Processing Systems}, 32, 2019.

\bibitem[Zeng and Nie(2021)]{zeng-nie-2021-investigation}
Yan Zeng and Jian-Yun Nie.
\newblock An investigation of suitability of pre-trained language models for
  dialogue generation {--} avoiding discrepancies.
\newblock In \emph{Findings of the Association for Computational Linguistics:
  ACL-IJCNLP 2021}, pages 4481--4494, Online, August 2021. Association for
  Computational Linguistics.
\newblock \doi{10.18653/v1/2021.findings-acl.393}.

\bibitem[Thrush et~al.(2022)Thrush, Jiang, Bartolo, Singh, Williams, Kiela, and
  Ross]{thrush2022winoground}
Tristan Thrush, Ryan Jiang, Max Bartolo, Amanpreet Singh, Adina Williams, Douwe
  Kiela, and Candace Ross.
\newblock Winoground: Probing vision and language models for visio-linguistic
  compositionality.
\newblock In \emph{Proceedings of the IEEE/CVF Conference on Computer Vision
  and Pattern Recognition}, pages 5238--5248, 2022.

\bibitem[Chung et~al.(2022)Chung, Hou, Longpre, Zoph, Tay, Fedus, Li, Wang,
  Dehghani, Brahma, et~al.]{chung2022scaling}
Hyung~Won Chung, Le~Hou, Shayne Longpre, Barret Zoph, Yi~Tay, William Fedus,
  Eric Li, Xuezhi Wang, Mostafa Dehghani, Siddhartha Brahma, et~al.
\newblock Scaling instruction-finetuned language models.
\newblock \emph{arXiv preprint arXiv:2210.11416}, 2022.

\bibitem[Bravo et~al.(2023)Bravo, Mittal, Ging, and Brox]{bravo2023open}
Maria~A Bravo, Sudhanshu Mittal, Simon Ging, and Thomas Brox.
\newblock Open-vocabulary attribute detection.
\newblock In \emph{Proceedings of the IEEE/CVF Conference on Computer Vision
  and Pattern Recognition}, pages 7041--7050, 2023.

\bibitem[Xu et~al.(2016)Xu, Mei, Yao, and Rui]{xu2016msr}
Jun Xu, Tao Mei, Ting Yao, and Yong Rui.
\newblock Msr-vtt: A large video description dataset for bridging video and
  language.
\newblock In \emph{Proceedings of the IEEE conference on computer vision and
  pattern recognition}, pages 5288--5296, 2016.

\bibitem[Xu et~al.(2017)Xu, Zhao, Xiao, Wu, Zhang, He, and Zhuang]{xu2017video}
Dejing Xu, Zhou Zhao, Jun Xiao, Fei Wu, Hanwang Zhang, Xiangnan He, and Yueting
  Zhuang.
\newblock Video question answering via gradually refined attention over
  appearance and motion.
\newblock In \emph{Proceedings of the 25th ACM international conference on
  Multimedia}, pages 1645--1653, 2017.

\bibitem[Young et~al.(2014)Young, Lai, Hodosh, and Hockenmaier]{young2014image}
Peter Young, Alice Lai, Micah Hodosh, and Julia Hockenmaier.
\newblock From image descriptions to visual denotations: New similarity metrics
  for semantic inference over event descriptions.
\newblock \emph{Transactions of the Association for Computational Linguistics},
  2:\penalty0 67--78, 2014.
\newblock \doi{10.1162/tacl_a_00166}.

\bibitem[Yoshikawa et~al.(2017)Yoshikawa, Shigeto, and
  Takeuchi]{yoshikawa2017stair}
Yuya Yoshikawa, Yutaro Shigeto, and Akikazu Takeuchi.
\newblock {STAIR} captions: Constructing a large-scale {J}apanese image caption
  dataset.
\newblock In \emph{Proceedings of the 55th Annual Meeting of the Association
  for Computational Linguistics (Volume 2: Short Papers)}, pages 417--421,
  Vancouver, Canada, 2017. Association for Computational Linguistics.
\newblock \doi{10.18653/v1/P17-2066}.

\bibitem[Li et~al.(2019{\natexlab{b}})Li, Xu, Wang, Lan, Jia, Yang, and
  Xu]{li2019coco}
Xirong Li, Chaoxi Xu, Xiaoxu Wang, Weiyu Lan, Zhengxiong Jia, Gang Yang, and
  Jieping Xu.
\newblock Coco-cn for cross-lingual image tagging, captioning, and retrieval.
\newblock \emph{IEEE Transactions on Multimedia}, 21\penalty0 (9):\penalty0
  2347--2360, 2019{\natexlab{b}}.

\bibitem[Bugliarello et~al.(2022)Bugliarello, Liu, Pfeiffer, Reddy, Elliott,
  Ponti, and Vuli{\'c}]{bugliarello2022iglue}
Emanuele Bugliarello, Fangyu Liu, Jonas Pfeiffer, Siva Reddy, Desmond Elliott,
  Edoardo~Maria Ponti, and Ivan Vuli{\'c}.
\newblock Iglue: A benchmark for transfer learning across modalities, tasks,
  and languages.
\newblock \emph{ArXiv preprint}, abs/2201.11732, 2022.

\end{thebibliography}

\appendix

\newpage

\section{Appendix}

\subsection{Pre-training Datasets}

As follows, we give some data filtering details. Since LAION and the video-text datasets are too large, we have filtered the datasets to speed up the pre-training. Specifically, for LAION, we use English data only. Following BLIP~\cite{li2022blip}, we remove an image if the shorter edge is smaller than 224 pixels. We also remove an image if the ratio of height/width or width/height is larger than 3. For video clip-text pairs, we remove a pair if the number of words is less than 2. Following previous work, we use CLIP score to filter video data. We sample a frame for a video clip and we calculate the CLIP score between the frame and the text. We remove a video clip-text pair if the score is less than 0.25. For image annotations of objects and regions, we remove a sample because of: 1) invalid annotations (e.g. negative values for bounding boxes or boxes being outside of the images); 2) boxes being too small (less than a patch); 3) highly overlapped text descriptions of regions ($>$ 75\%), etc. For an object annotation, if it contains an object attribute, e.g. color, we concatenate the attribute with the object label as the text description. Moreover, some images in the OpenImages dataset contain relationship annotations, indicating pairs of objects in particular relations (e.g. "woman playing guitar", "beer on table"), object properties (e.g. "table is wooden"), and human actions (e.g. "woman is jumping"). We also utilize this part of data.

\subsection{Implementation Details}

\baby is pre-trained at image resolution of $224\times224$ using $16\times16$ patch size. Though, as indicated in previous work such as OFA~\cite{wang2022ofa} and CoCa~\cite{yu2022coca}, increasing resolution will improve model performance, we keep it small to accelerate pre-training. Besides, we apply mixed precision for training. For text input, we set the maximum number of tokens to 30. To further speed up pre-training with large-scale data, we divide the training process into two steps. First, we train \baby with large-scale image-text pairs. Then, we further train \baby on video-text pairs and the 4M dataset. The reason behind this is that training on video data is slow. Because of it, we randomly sample only three frames for a video clip in pre-training. We mix all types of data in a training batch, and thus for each training iteration, we optimize the model by multi-grained aligning and multi-grained localization simultaneously.

With 4M data, we pre-train \babyB for 500K steps with a batch size of 1024 on 8 A100 and \babyL for 250K steps on 16 A100, which takes $\sim1$ week. The learning rate of \babyB is warmed-up to $1e^{-4}$ in the first 2500 steps and decayed following a linear schedule. The learning rate is $5e^{-5}$ for \babyL. With large-scale data, training \baby takes 2-3 weeks on 32 A100 for the base model and 64 A100 for the large model.

\subsection{Ablation Study}
To ensure a fair comparison, all compared model variants are trained on 4M images for 100K steps. Following previous studies, we have shortened the training steps to compare different ablated variants more efficiently. We evaluate model performance on image-text retrieval (Recall@1), visual question answering, visual grounding, and zero-shot open-vocabulary attribute detection. It is worth noting that VQA has a large train and test set, which means that even a relatively small difference in performance is worth considering.

\begin{table}[h]
\centering
\resizebox{0.7\columnwidth}{!}{%
\begin{tabular}{lcccccc}
\toprule
\multirow{2}{*}{ } & \multicolumn{2}{c}{\bf Flickr30K} & \bf VQA & \bf MSRVTT & \multicolumn{2}{c}{\bf Video-QA}\\
 & TR & IR & test-dev & IR & MSRVTT & MSVD \\
\midrule
w/ avg pool (ours) & \bf 98.5 & \bf 90.4 & \bf 80.4 & \bf 47.6 & \bf 45.0 & \bf 52.8 \\

w/ temporal attn & 98.2 & 89.6 & 80.0 & 45.6 & 44.4 & 52.1 \\

\bottomrule
\end{tabular}
}
\vspace{0.2cm}

\caption{\textbf{Ablation study} of temporal modeling methods. 
}
\label{tbl:ab_video}
\end{table}

Additionally, we investigate whether better temporal modeling could further improve video understanding capabilities while maintaining good image understanding, as presented in Table~\ref{tbl:ab_video}. We use an established method that adds temporal attention in ViT. The experimental results on image/video-text retrieval and image/video VQA show that simply averaging the features of each frame achieves better performances on all tasks. We suppose that our approach is more unified in modeling both image and video features, and thus strong image understanding capability is better transferred to video understanding.

\subsection{Qualitative Study of Multi-Grained Alignments}

\begin{figure*}[ht]
\begin{center}
\centerline{\includegraphics[width=1\columnwidth]{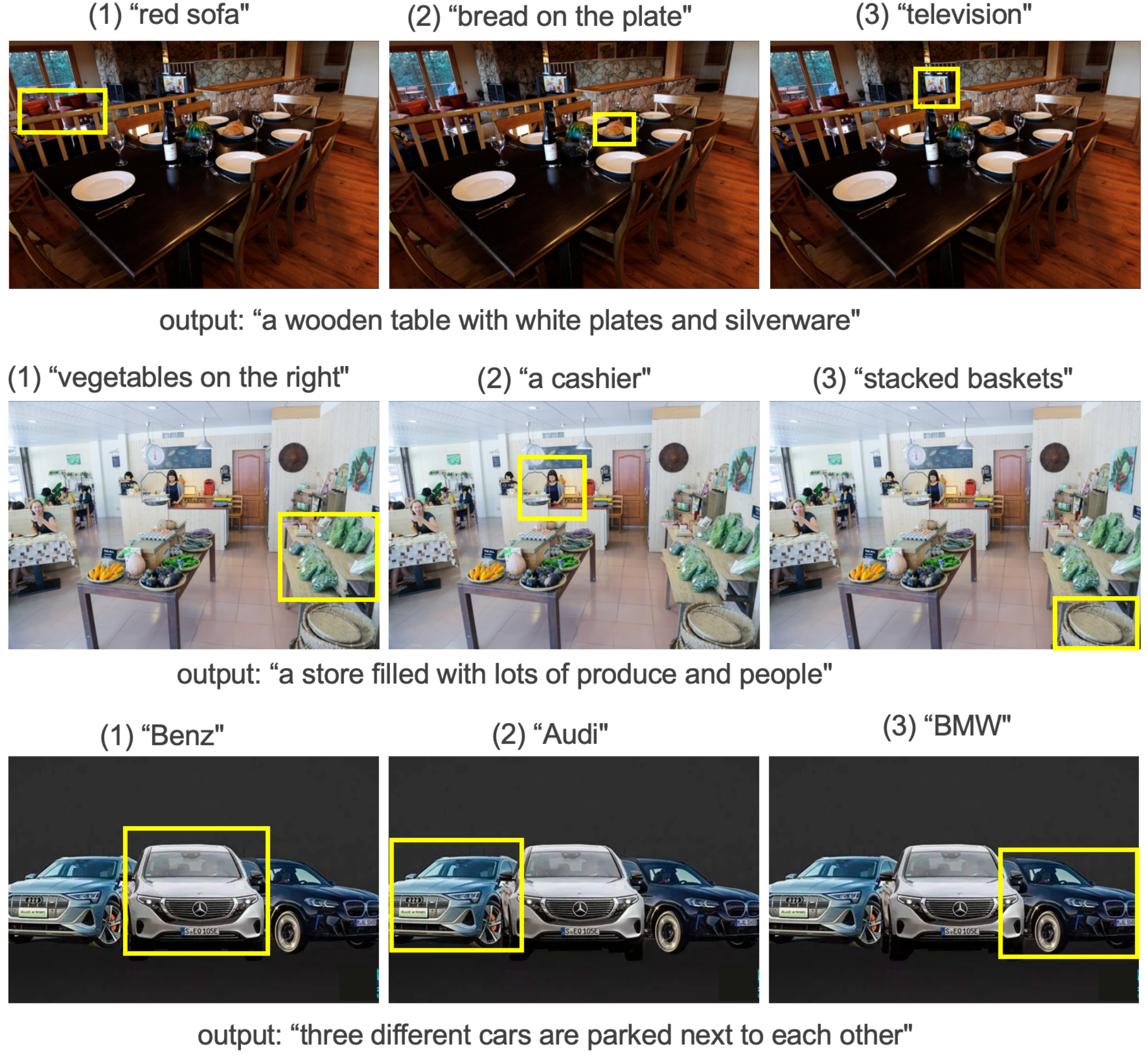}}
\caption{Visualization of \baby generating captions for images and locating visual concepts given manual input descriptions. }
\label{app:example1}
\end{center}
\end{figure*}

\begin{figure*}[ht]
\begin{center}
\centerline{\includegraphics[width=1\columnwidth]{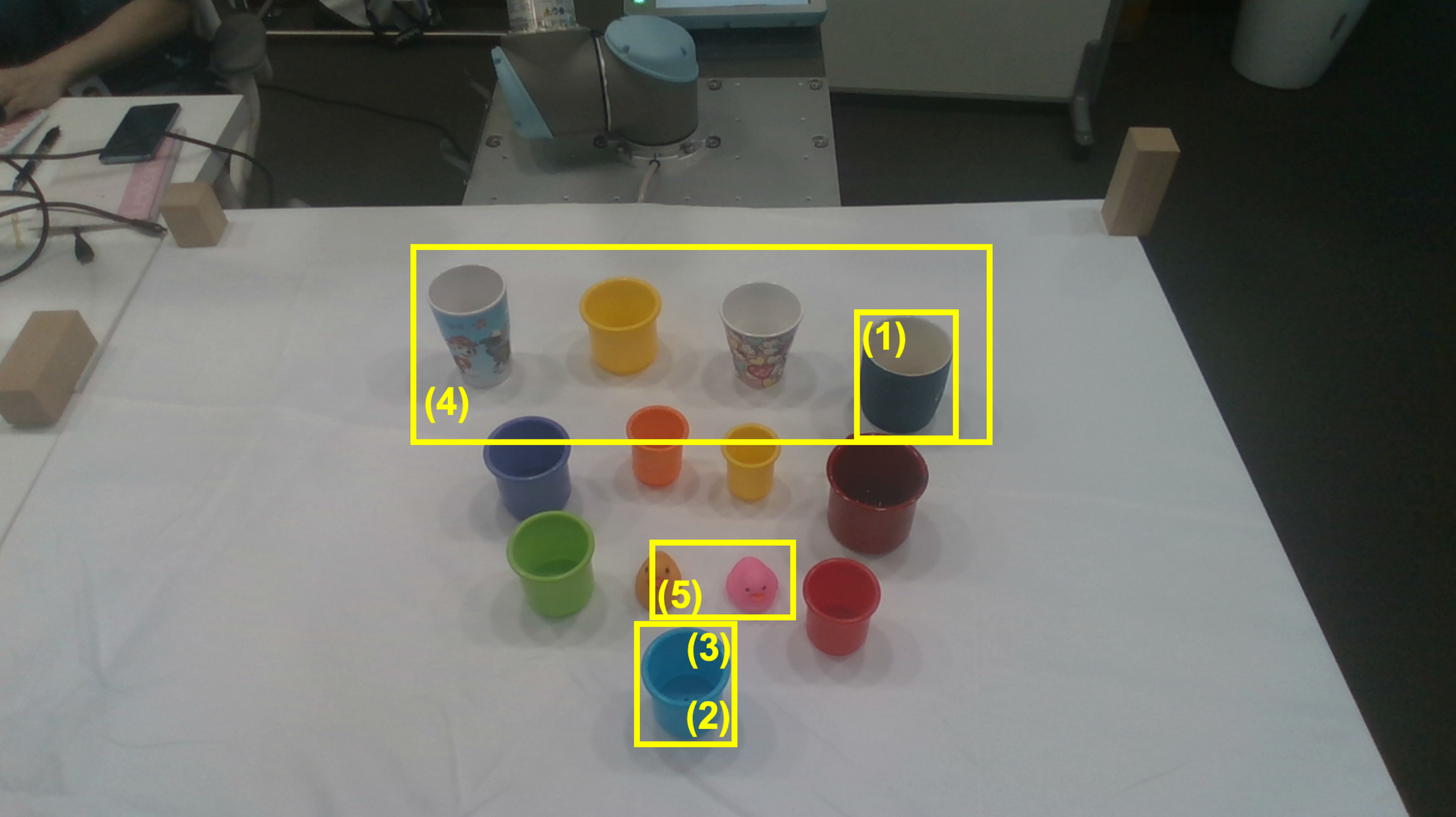}}
\caption{Visualization of \baby locating visual concepts in robot grasping scene given text descriptions: 1) ``deep blue cup''; 2) ``light blue cup''; 3) ``blue cup at the bottom''; 4) ``four cups at the top''; 5) ``two small ducks''. }
\label{app:g1}
\end{center}
\end{figure*}

\begin{figure*}[ht]
\begin{center}
\centerline{\includegraphics[width=1\columnwidth]{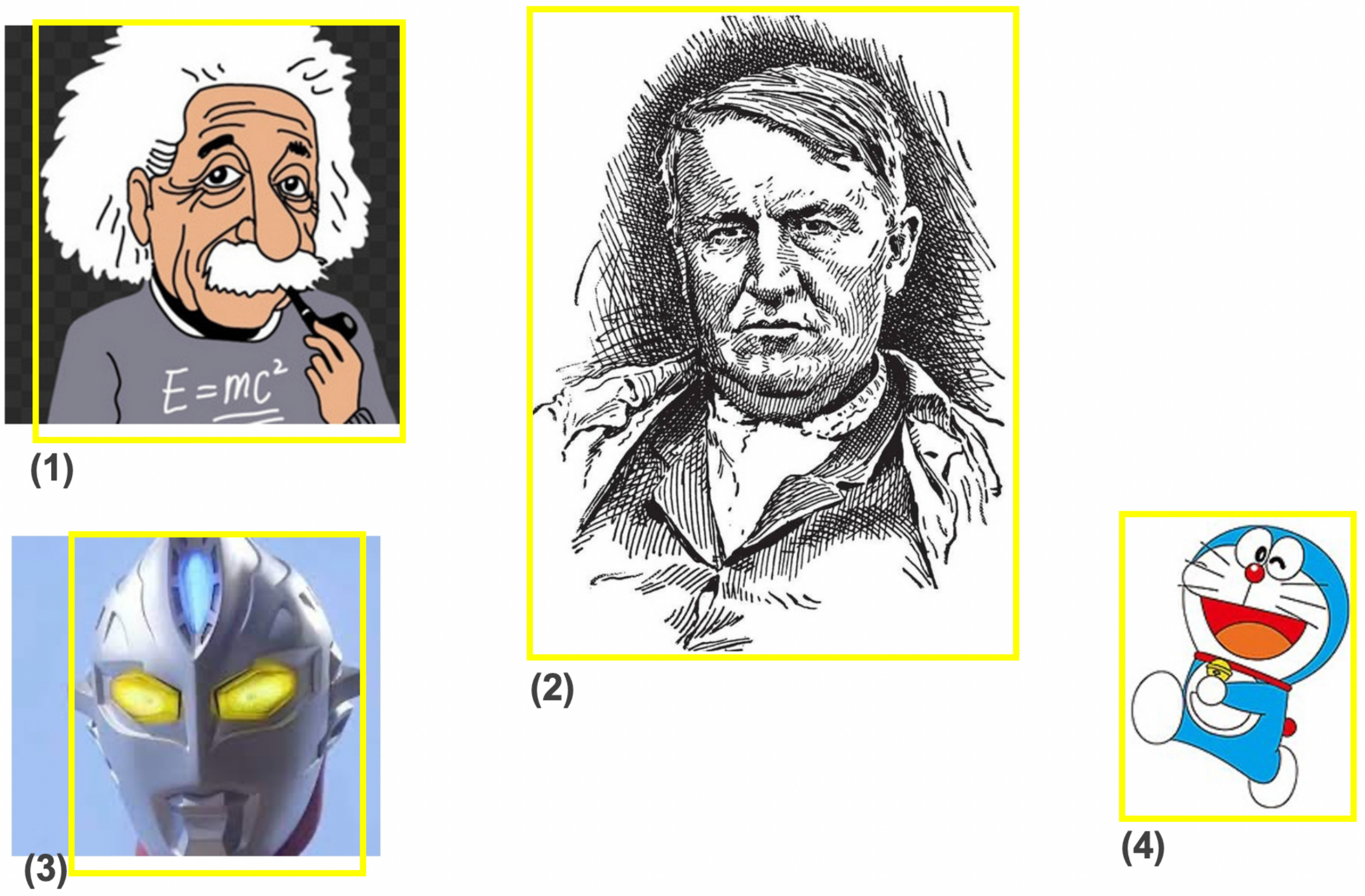}}
\caption{Visualization of \baby locating celebrities and cartoon characters given text descriptions: 1) ``Albert Einstein''; 2)``Edison''; 3)``Ultraman''; 4)``Doraemon''. }
\label{app:g2}
\end{center}
\end{figure*}

\begin{figure*}[ht]
\begin{center}
\centerline{\includegraphics[width=1\columnwidth]{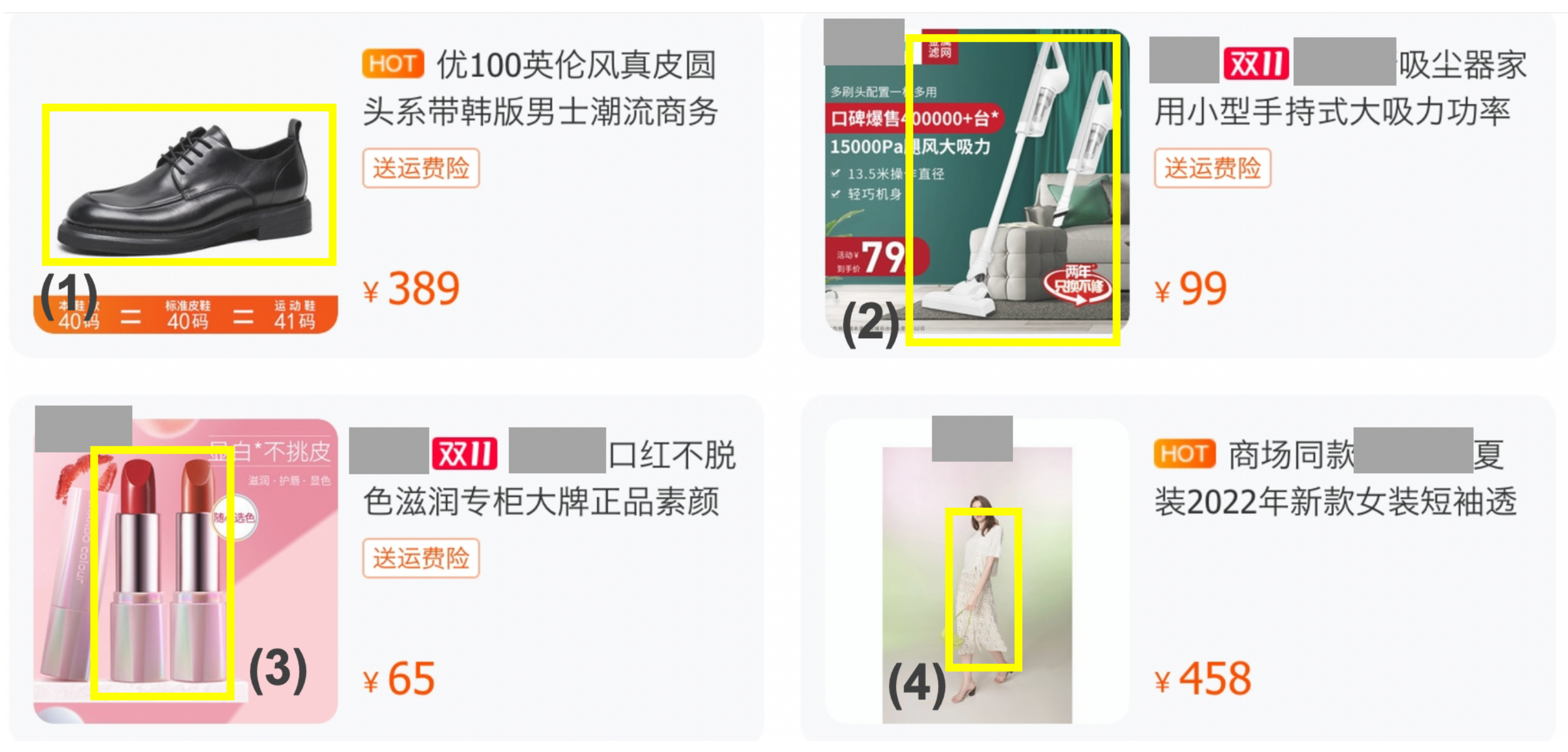}}
\caption{Visualization of \baby locating objects in an image from an e-commerce website in China. The input text descriptions are: 1) ``shoes''; 2)``vacuum cleaner''; 3)``lipstick''; 4)``dress''. }
\label{app:g3}
\end{center}
\end{figure*}

\begin{figure*}[ht]
\begin{center}
\centerline{\includegraphics[width=1\columnwidth]{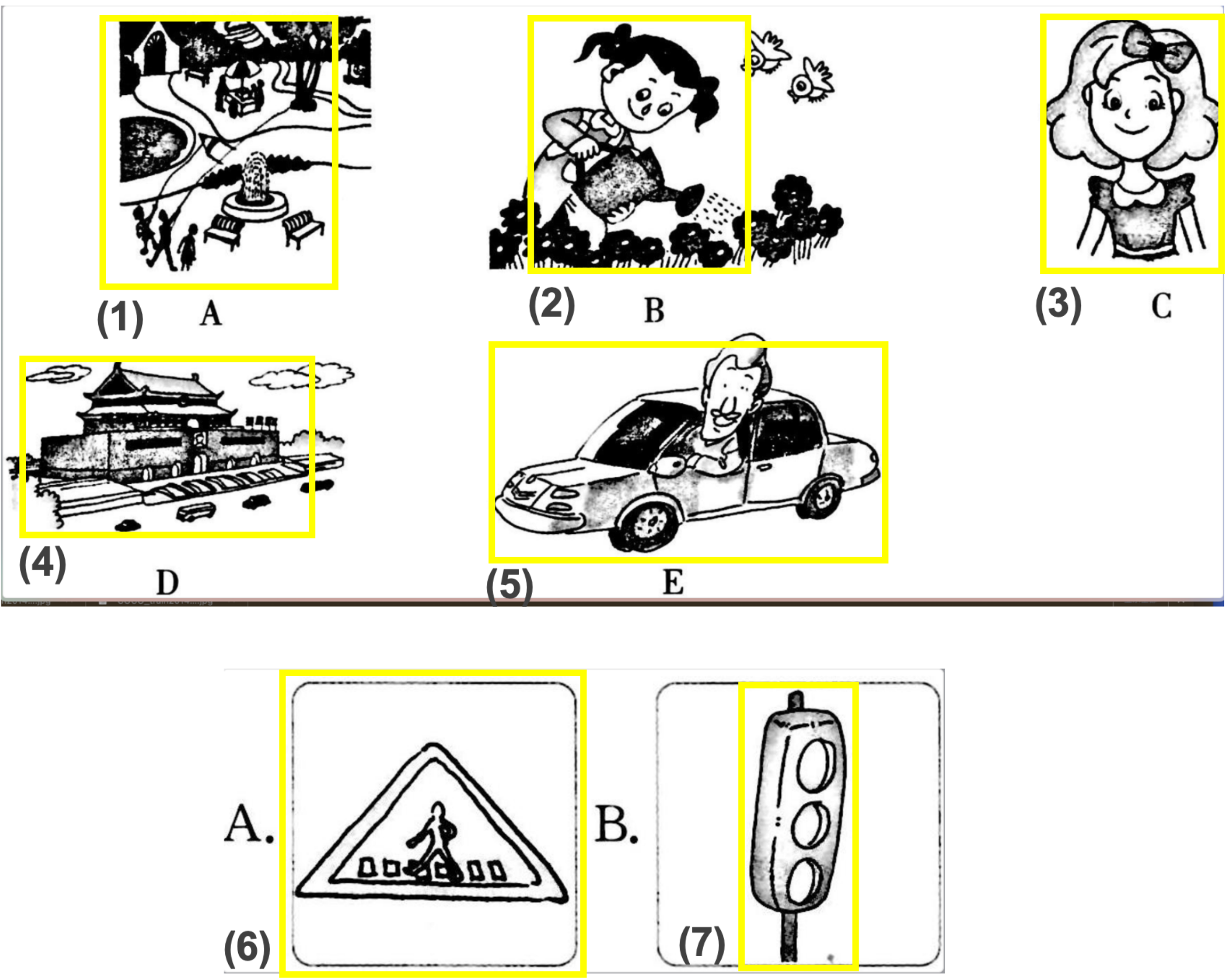}}
\caption{Visualization of \baby locating visual concepts in an image from children's textbooks. The input text descriptions are: 1) ``flying kites in the park''; 2) ``watering flowers''; 3) ``well dressed girl''; 4) ``Tiananmen Tower''; 5) ``drive to work''; 6) ``sign''; 7) ``traffic lights''. }
\label{app:g4}
\end{center}
\end{figure*}

We provide a qualitative study of what vision language alignments have been learned by \baby. To this end, we ask \baby to generate image captions or to locate visual concepts. We visualize the results in Figure~\ref{app:example1}, where the first two images are from the in-domain COCO dataset. We find that \baby can capture small objects in the background or objects which have been partly masked. We also choose out-of-domain images for evaluation. As shown in Figure~\ref{app:g1}, Figure~\ref{app:g2}, Figure~\ref{app:g3}, and Figure~\ref{app:g4}, \baby can recognize many visual concepts from different domains.

\end{document}